\documentclass[journal]{IEEEtran}
\usepackage{times}
\usepackage{epsfig}
\usepackage{graphicx}
\usepackage{amsmath}
\usepackage{amssymb}
\usepackage{multirow}
\usepackage{enumerate}
\usepackage{booktabs}

\usepackage[pagebackref=true,breaklinks=true,letterpaper=true,colorlinks,bookmarks=false]{hyperref}

\newcommand{\etal}{\textit{et al}.}
\newcommand{\ie}{\textit{i}.\textit{e}.}
\newcommand{\eg}{\textit{e}.\textit{g}.}

\ifCLASSINFOpdf
\else
\fi
\newcommand{\titlename}{Deep Rank-Consistent Pyramid Model for Enhanced Crowd Counting}
\newcommand{\methodname}{DREAM}
\newcommand{\gao}[1]{\textcolor{black}{#1}}
\newcommand{\re}[1]{\textcolor{black}{#1}}
\newcommand{\redd}[1]{\textcolor{black}{#1}}

\begin{document}
%
\title{\titlename}
%
%
%

\author{
        Jiaqi~Gao,~\IEEEmembership{Graduate Student Member,~IEEE},
        Zhizhong~Huang,~\IEEEmembership{Graduate Student Member,~IEEE},
        Yiming~Lei,~\IEEEmembership{Member,~IEEE},
        Hongming Shan,~\IEEEmembership{Senior Member,~IEEE},
        James Z. Wang,~\IEEEmembership{Senior Member,~IEEE},\\
        Fei-Yue Wang$^\dag$,~\IEEEmembership{Fellow,~IEEE,}
        Junping Zhang$^\dag$,~\IEEEmembership{Senior Member,~IEEE}
\thanks{This work was supported in part by the National Natural Science Foundation of China (Nos. 62176059, 62101136, T2192933, 61533019), in part by Shanghai Municipal Science and Technology Major Project under Grant 2018SHZDZX01, in part by the ZJ Lab, the Shanghai Center for Brain Science and Brain-inspired Technology. J. Z. Wang was supported by The Pennsylvania State University.}
\thanks{J. Gao, Z. Huang, Y. Lei, and J. Zhang are with the Shanghai Key Laboratory of Intelligent Information Processing, School of Computer Science, Fudan University, Shanghai 200433, China (e-mail: \{jqgao20, zzhuang19, ymlei17, jpzhang\}@fudan.edu.cn).}
\thanks{H. Shan is with the Institute of Science and Technology for Brain-inspired
Intelligence and MOE Frontiers Center for Brain Science, Fudan University,
Shanghai 200433, China, and also with the Shanghai Center for Brain
Science and Brain-Inspired Technology, Shanghai 201210, China. (Email:
hmshan@fudan.edu.cn)}
\thanks{J. Z. Wang is with the College of Information Sciences and Technology,
The Pennsylvania State University, University Park, PA 16802 USA. (e-mail:
jwang@ist.psu.edu).}
\thanks{F. Wang is with the State Key Laboratory of Management and Control for Complex Systems, Institute of Automation, Chinese Academy of Sciences,
Beijing 100190, China, also with the Institute of Systems Engineering,
Macau University of Science and Technology, Macau, China, and also with
Qingdao Academy of Intelligent Industries, Qingdao 266109, China (e-mail:
feiyue.wang@ia.ac.cn).}
\thanks{\dag~Co-corresponding Authors.}

}

%
%

\markboth{}%
{J. Gao \MakeLowercase{\textit{et al.}}: \titlename}
%



\maketitle

\begin{abstract}
Most conventional crowd counting methods utilize a fully-supervised learning framework to establish a mapping between scene images and crowd density maps. \re{They} usually \re{rely} on a large quantity of costly and time-intensive pixel-level annotations for training supervision. One way to \re{mitigate the intensive labeling effort and improve counting accuracy} is to \re{leverage large amounts of unlabeled images. This is attributed to the inherent self-structural information and rank consistency within a single image, offering additional qualitative relation supervision during training.} Contrary to earlier methods that utilized the {\re{rank relations}} at the original image level, we explore \re{such rank-consistency relation} within the latent feature spaces. \re{This approach enables the incorporation of numerous pyramid partial orders, strengthening the model representation capability. A notable advantage is that it can also increase the utilization ratio of unlabeled samples.} Specifically, we propose a \textbf{D}eep \textbf{R}ank-consist\textbf{E}nt pyr\textbf{A}mid \textbf{M}odel~(\methodname), which \re{makes full use of rank consistency across} coarse-to-fine pyramid features in latent spaces for enhanced crowd counting with massive unlabeled images. In addition, we have collected a new unlabeled crowd counting dataset, FUDAN-UCC, comprising 4,000 images for training purposes. Extensive experiments on four benchmark datasets, namely UCF-QNRF, ShanghaiTech PartA and PartB, and UCF-CC-50, show the effectiveness of our method compared with previous semi-supervised methods. \gao{The codes are available at \url{https://github.com/bridgeqiqi/DREAM}}.

\end{abstract}


\IEEEpeerreviewmaketitle

\section{Introduction}
\label{section: introduction}
\IEEEPARstart{C}{rowd} counting has broad applications in traffic management, public safety surveillance, smart city planning such as preventing stampedes, and estimating participation in rallies or parades. During health crises, such as the COVID-19 pandemic, effective crowd counting can help authorities determine whether social distancing measures are feasible in specific public areas. The primary goal of crowd counting is to estimate the number of people present in a given image or video sequence, especially in crowded scenes. Although crowd counting has been an active research area in recent years, the task remains a challenge due to the influence of many extrinsic factors, including occlusion, illumination, head size variations, diverse perspectives, and uneven distribution of crowds. \gao{It may also suffer from the significant costs in terms of time and resources in labeling dense scenes.}

\begin{figure}[t]
   \centering
   \includegraphics[width=0.95\linewidth]{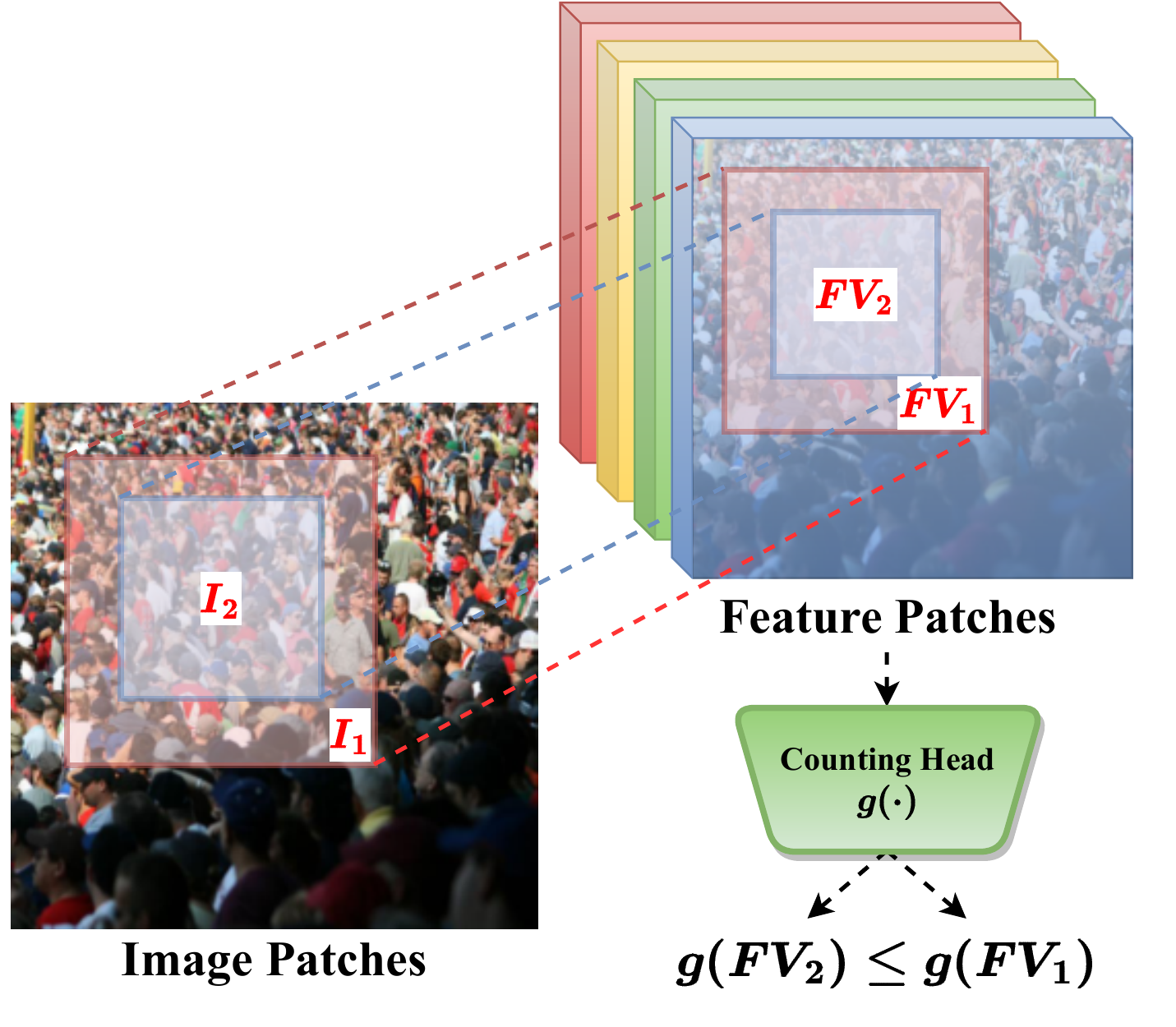}
   \caption{\gao{The motivation of our work.} $FV_1$ and $FV_2$ are the rank-consistent feature patches cropped on feature maps in the latent space. Their corresponding regions in the input space are $I_1$ and $I_2$, respectively, according to the receptive field. Thus, it should be guaranteed that the output counts $g(FV_2)$ predicted by larger patches in the latent space would be \gao{no greater than} the counts $g(FV_1)$ predicted by their sub-patches in the same feature {map}.}
   \label{figures: observation}
\end{figure}

Earlier counting methods were based on detection techniques~\cite{leibe2005pedestrian,dollar2011pedestrian}. They mainly focused on designing robust human-pose or human-body detectors to count pedestrians in scenes using a sliding-window template matching approach~\cite{dollar2011pedestrian}. {\re{This process is both computationally expensive and time-consuming. Meanwhile, its accuracy to a great extent depends on detector performance, which may suffer from the occlusion, illumination, and scale variation issues.}} As an alternative, some researchers treated crowd counting as a regression task~\cite{chan2009bayesian,ryan2009crowd,chen2012feature} by learning a mapping from the original image to the final count number\re{, which can accelerate both the training and inference process}.

Both paradigms have shown proficiency in sparse scenes. Even with accuracy improvements by several state-of-the-art object detection methods~\cite{ren2015faster,redmon2016you,redmon2018yolov3,liu2016ssd, jiao2021new, cao2021hierarchical, wu2022enhanced,li2022dagcn} with the help of extracting multi-scale features, performance can be hindered due to occlusion~\cite{tan2022mhsa, miao2021identifying}, congested scenes~\cite{wang2022review} and tiny head sizes for individuals situated far from the camera's perspective. Benefiting from the strong representation learning ability of convolutional neural networks (CNN), in recent years, CNN-based methods~\cite{zhang2015cross,zhang2016single,sindagi2017generating,babu2017switching,boominathan2016crowdnet,liu2019adcrowdnet,li2018csrnet,liu2019context,tian2019padnet,cao2018scale,chen2019scale,jiang2019learning,dong2022clrnet, gao2021domain} predict a density map from a still image for crowd counting. \re{The CNN-based density map estimation paradigm helps localize the people more accurately because density maps capture more spatial information of people distribution and provide pixel-wise supervision. } Thus, counting performance has been significantly improved. Accurate visual localization often plays a vital role in other related crowd analysis and surveillance tasks, such as object segmentation~\cite{zhou2021self,li2022self,xian2022location, gu2022pixel}, tracking~\cite{ge2020cascaded,yu2022learning,guo2023multi}, and context modeling in autonomous driving~\cite{predhumeau2023pedestrian,zhou2023multi, guan2023autonomous, huang2023cost}.

{\re{Despite the promising results of most existing CNN-based methods on several public datasets~\cite{idrees2013multi,zhang2016single,idrees2018composition,wang2020nwpu}, the challenge of requiring vast amounts of data remains. The labor-intensive and expensive labeling process results in relatively small labeled training datasets, especially for highly crowded scenes.}} For instance, it takes over 30 minutes for an adult to annotate a single image with an immensely congested scene containing over 2,000 people. \re{Thus, the main goal of this paper is to investigate methods to leverage limited labeled images and large amounts of unlabeled ones, aiming to boost the ultimate counting accuracy while alleviating the extensive data demand and reducing labeling costs.}
Synthetic images generated by game environments~\cite{wang2019learning, zhang2022game} may serve as a potential solution to address the labeled data shortage in the real world. Models can be initially trained on these synthetic images in a supervised way and then be fine-tuned to real-world datasets. However, quite a few physical differences between the characters in the game and pedestrians in the real world must be handled. Other 
solutions~\cite{liu2018leveraging,sindagieccv2020learning,yanliusemieccv2020} are to leverage limited labeled images and abundant unlabeled images for semi-supervised crowd counting. Specifically, auxiliary tasks, such as the Gaussian process~\cite{sindagieccv2020learning} and 
segmentation surrogate tasks~\cite{yanliusemieccv2020}, are used to generate pseudo-labels for unlabeled images to assist the feature extractor to learn more robust feature representations and help the crowd counter learn a more discriminative decoder. The L2R method~\cite{liu2018leveraging} exploits the structural rank consistency of unlabeled images at the image level to assist the counter in predicting more accurate density maps. However, the generation quality of pseudo labels 
depends largely upon the model capacity and representation learning ability. Surrogate tasks may introduce extra computational costs and parameters. Meanwhile, simple constraints among different image scales are limited to reduce the estimation errors.

In this paper, we propose a \textbf{D}eep \textbf{R}ank-consist\textbf{E}nt pyr\textbf{A}mid \textbf{M}odel (\methodname), a novel semi-supervised approach that can leverage more partial orders of pyramid rank-consistent features from coarse-to-fine stages among unlabeled images to predict more accurate density maps from limited labeled samples. \re{Each pixel within any intermediate feature map can represent the high-level semantic information of its corresponding region (receptive field) in the original input image.} 
These intermediate features taken from unlabeled images \re{should be rank-consistent, driving} the model to learn more general and robust representations. {\redd{Different from the L2R model~\cite{liu2019exploiting}, we exploit rank consistency among unlabeled images in the pyramid feature levels. Meanwhile, our method can further improve the sample utilization efficiency of unlabeled samples because the number of rank-consistent pairs of our method is generated from the features at three different scales, which are at least three times the ones on the images in L2R. The main difference between our method and L2R~\cite{liu2019exploiting} is that our method performs the ranking at the pyramid feature level, in contrast to the single image level in~\cite{liu2019exploiting}. 
This design can encourage the counting model to learn better counting-specific features by maintaining ranking consistency at different feature levels.}} To further facilitate the model's learning from limited labeled images, we propose a new \textbf{U}nlabeled \textbf{C}rowd \textbf{C}ounting dataset, \textbf{FUDAN-UCC}. Comprising 4,000 images from the image search engine GettyImages\footnote{https://www.gettyimages.com/}, this collection will serve as the unlabeled dataset throughout the training process.

Our \textbf{main contributions} can be summarized as follows.
\begin{itemize}
    \item \gao{We propose a semi-supervised crowd counting framework that exploits the rank-consistent pyramid features among unlabeled images to enhance counting accuracy. With the help of the novel coarse-to-fine feature margin ranking loss, the model could utilize partial orders and rank-consistent information among unlabeled images to estimate the counts more accurately with limited labeled images. The proposed coarse-to-fine ranking loss on the feature level is both intuitive and straightforward to implement.}
    \item We have constructed a large unlabeled crowd counting dataset, FUDAN-UCC, from the Internet. With 4,000 high-resolution images of densely populated scenes, the dataset paves the way for robust, unbiased comparisons among semi-supervised crowd counting methods.
    \item Extensive experiments on several benchmark crowd counting datasets show the effectiveness of our proposed \methodname model.
\end{itemize}

\section{Related Work}
\label{section: related work}

\subsection{Conventional Counting Methods}
\textbf{Detection-based Methods:}
Using object detection for crowd counting~\cite{zhang2020kgsnet} is an intuitive approach. In earlier studies, detectors were trained using classical hand-crafted features such as SIFT, HOG, and edge features extracted from the part or whole of human bodies. Researchers~\cite{dalal2005histograms,leibe2005pedestrian, enzweiler2008monocular,tuzel2008pedestrian} extracted the general features from the entire body to train  classifiers using algorithms such as SVM, boosting, and random forest. However, they achieved limited performance, especially in scenes with significant occlusions. In contrast, body-part features~\cite{felzenszwalb2009object,wu2005detection}, \eg, heads and shoulders, improved the accuracy to some extent. Nevertheless, counting-by-detection methods primarily excel in sparse scenes, given their sensitivity to heavy occlusions and density variations.

\textbf{Regression-based Methods:}
Regression-based approaches focus on enhancing the capacity to estimate global counts in crowd counting. They typically seek to learn a mapping function from both local and global features to the overall count~\cite{chen2012feature,chan2011counting, chan2009bayesian}. The process generally comprises two steps: i) Extracting useful features, including foreground features,
textures, corners, histogram oriented gradients (HoG), and local binary patterns (LBP); and ii) Training a regression model such as linear regression, ridge regression, Bayesian Poisson regression~\cite{chan2009bayesian}, and Gaussian process regression~\cite{sindagieccv2020learning} based on features extracted from step i).

Nevertheless, these conventional counting approaches mainly rely on hand-crafted features and may falter in extremely crowded scenes. Meanwhile, by overlooking pedestrian distribution, their performance remain constrained. To better learn spatial distribution information of individuals within a scene, Lempitsky \etal ~\cite{lempitsky2010learning} proposed to predict a density map instead of regressing a scalar for crowd counting. A density map can reflect the distribution of people approximately and its integral is equal to the number of people in a given image.

\subsection{CNN-based Methods}
\textbf{Supervised Methods:} Most off-the-shelf deep learning-based methods are built upon stacks of convolution operations to regress crowd counts or density maps. One density map could not only predict the potential location of each person but also explain the overview of spatial distribution.
More specifically, Wang \etal~\cite{wang2015deep} used an AlexNet-like architecture to predict the number of people in highly crowded scenes. Considering the perspective information, Zhang \etal~\cite{zhang2015cross} achieved better counting for unseen images in cross-scene scenarios. Zhang \etal~\cite{zhang2016single} further proposed an MCNN architecture, which contains multi-column convolutional layers with different kernel sizes, to resolve the scale variation issue for crowd counting. Following this, several multi-column models were developed. Sam \etal~\cite{babu2017switching} designed a switchable network for training patches within different density levels. Sindagi \etal ~\cite{sindagi2017generating} introduced both local and global contextual information to 
assist the network in generating high-quality density maps.
Observing that features extracted from different columns normally have similar and redundant patterns, Li \etal~\cite{li2018csrnet} proposed a deeper, single-column network within dilated convolutions to address the issues and achieve better performance. Jiang \etal~\cite{jiang2019learning} fused features from different layers by simple concatenations in CNN to obtain a multi-scale feature map representation.
Besides, Liu \etal~\cite{liu2019context} combined scale-aware contextual features with perspective maps to estimate density maps, while Cao \etal~\cite{cao2018scale} and Chen \etal ~\cite{chen2019scale} designed a scale aggregation network and a scale pyramid network to tackle the scale variation problems. {\redd{Chen \etal~\cite{chen2023crowd} proposed a multi-scale spatial guided perception aggregation network~(MGANet) to deal with the dramatic scale variation issues in a single image}}. Jiang \etal~\cite{jiang2020attention} obtained segmentation masks according to regions of different density levels and introduced a scaling factor to jointly estimate people counts.
Yang \etal~\cite{yang2020reverse} uniformly warped the input images to normalize head sizes at different locations to the same scale through perspective transformations. Furthermore, to correct small errors of ground truth caused by the empirically chosen parameter of head scale $\sigma$, Wan \etal~\cite{wan2019adaptive,wan2020kernel} utilized the kernel-based density map to refine the final density map. {\redd{Zhu \etal~\cite{zhu2023confusion} designed an end-to-end confusion region discriminating and erasing network~(CDENet) to address the incorrect estimation problems among confusion regions.}} Bai \etal~\cite{bai2020adaptive} self-corrected the density map by EM algorithm. 

Adversarial networks~\cite{shen2018crowd,yang2018multi}
have also been used in crowd counting to generate high-quality density maps.
Sindagi \etal~\cite{sindagi2019multi} fused multi-level bottom-top and top-bottom features to
address the scale variation problems in crowd counting.
Xiong \etal~\cite{xiong2019open} divided feature maps into several grids and counted them hierarchically. A few deep reinforcement learning approaches~\cite{lu2022counting, xiao2023sampled} have also been introduced to enhance the crowd counting task. For example, Le \etal~\cite{lu2022counting} dealt with crowd counting from a sequence decision-making perspective, \ie weighing densities.
Ma \etal~\cite{ma2019bayesian} proposed a Bayesian loss to learn an expectation of people distribution by using point supervisions instead of generating density maps. 
{\re{
To estimate the changes of head sizes in a single image more accurately, Lian \etal~\cite{lian2021locating, lian2019density} leveraged a depth prior information and designed a depth-adaptive kernel to generate high-fidelity ground truth density maps for better training.
}}

\textbf{Semi-Supervised/Weakly-Supervised Methods:} Fully-supervised counting methods usually require massive pixel-level annotations, which can be prohibitively costly. 
A synthetic dataset~\cite{wang2019learning} constructed by the GTA5 game environment may solve the data-hungry issue. Another way is to leverage the vast repositories of unlabeled crowd images to assist the model in learning task-specific representations for more accurate crowd counting. For instance, Change \etal~\cite{change2013semi} implemented a unified active and semi-supervised regression framework to exploit the manifold structure of images. There are two approaches in semi-supervised settings: a) Generating a set of reliable pseudo labels~\cite{wang2021neuron}
for unlabeled images and then tune the model in a supervised way. Sindagi~\etal~\cite{sindagieccv2020learning} and Liu \etal ~\cite{yanliusemieccv2020} employed the Gaussian process method
and surrogate segmentation tasks, respectively, to generate the pseudo-labels for unlabeled images. b) Exploiting self-structural information and constructing an unsupervised loss among unlabeled images as an auxiliary loss to optimize the model. Liu \etal ~\cite{liu2018leveraging} leveraged more unlabeled data collected from the Internet and constructed a rank margin loss to optimize the model. Sam \etal ~\cite{sam2019almost} constructed an unsupervised reconstruction loss to learn useful features from unlabeled images and then trained the counter using labeled images. Parameters in the front layers are frozen when training the subsequent layers. 

Utilizing rank-consistent information among unlabeled images is crucial when counting is based on limited labeled samples. Although Liu \etal ~\cite{liu2018leveraging} claimed that such ordinal relations among unlabeled images in the input space are effective, these constraints are actually insufficient. In this paper, we explore such relations at different feature levels because features in deep layers of CNN contain more semantic information that is closer to densities and distributions. By the definition of receptive field, relative positions remain consistent through stacks of convolution operations. In other words, a feature patch within an intermediate feature map is supposed to correspond to a specific sub-region of the given input image~(See Fig.~\ref{figures: observation}). Therefore, the output predicted by one feature patch, $g(FV_2)$, should be no greater than that of its sub-patch, $g(FV_1)$. We regard these relations as `partial orders.' In our research, we exploit these partial orders across multiple intermediate layers and construct a coarse-to-fine feature pyramid margin rank loss to assist the model in learning from unlabeled images. In this case, our proposed {\re{\methodname~}}model can increase the utilization ratio of unlabeled images which are at least 3 times that of L2R~\cite{liu2018leveraging}.

\subsection{Learning to Rank}
{Learning to rank}~\cite{liu2009learning} aims to rank the items according to their relevance to a given query, which is a critical research topic in many applications, including information retrieval~\cite{li2022learning, ghanbari2022learning}, recommender systems~\cite{chen2023set, zehlike2022fairness, wang2023skellam}, confidence estimation~\cite{li2022confidence}, image retrieval~\cite{datta2008image, li2023gaitcotr, li2023motion}, and image quality assessment~\cite{ zagoruyko2015learning, faigenbaum2022image, liu2017rankiqa}.
In information retrieval~\cite{li2022learning, ghanbari2022learning}, a ranking function computes and assigns scores to texts, documents, or images, and then sorts them in descending order to facilitate retrieval. In computer vision tasks, Zagoruyko~\etal~\cite{zagoruyko2015learning} analyzed the feature similarity with different CNN architectures by comparing the image patches. Faigenbaum-Golovin~ \etal~\cite{faigenbaum2022image} ranked the pairwise image distortion level to assess the image quality. Additionally, Liu~\etal~\cite{liu2017rankiqa} used learning from rankings as a data augmentation technique to train a large neural network that is prepared for fine-tuning.

Our approach has a similar idea to the `learning to rank' methodology, specifically in the ranking of different feature pairs. However, there are two main distinctions. \emph{First}, our ranking process is unsupervised; there are no human annotations during the ranking process, which contrasts with the common supervised learning approach used in learning to rank. \emph{Second}, our method ranks different feature pairs with the human densities through an intuitive concept of rank consistency between the sub-regions of features, instead of relying on a relevance score in learning to rank.

\section{Methods}
\label{section: method}

\begin{figure}[ht]
   \centering
   \includegraphics[width=\linewidth]{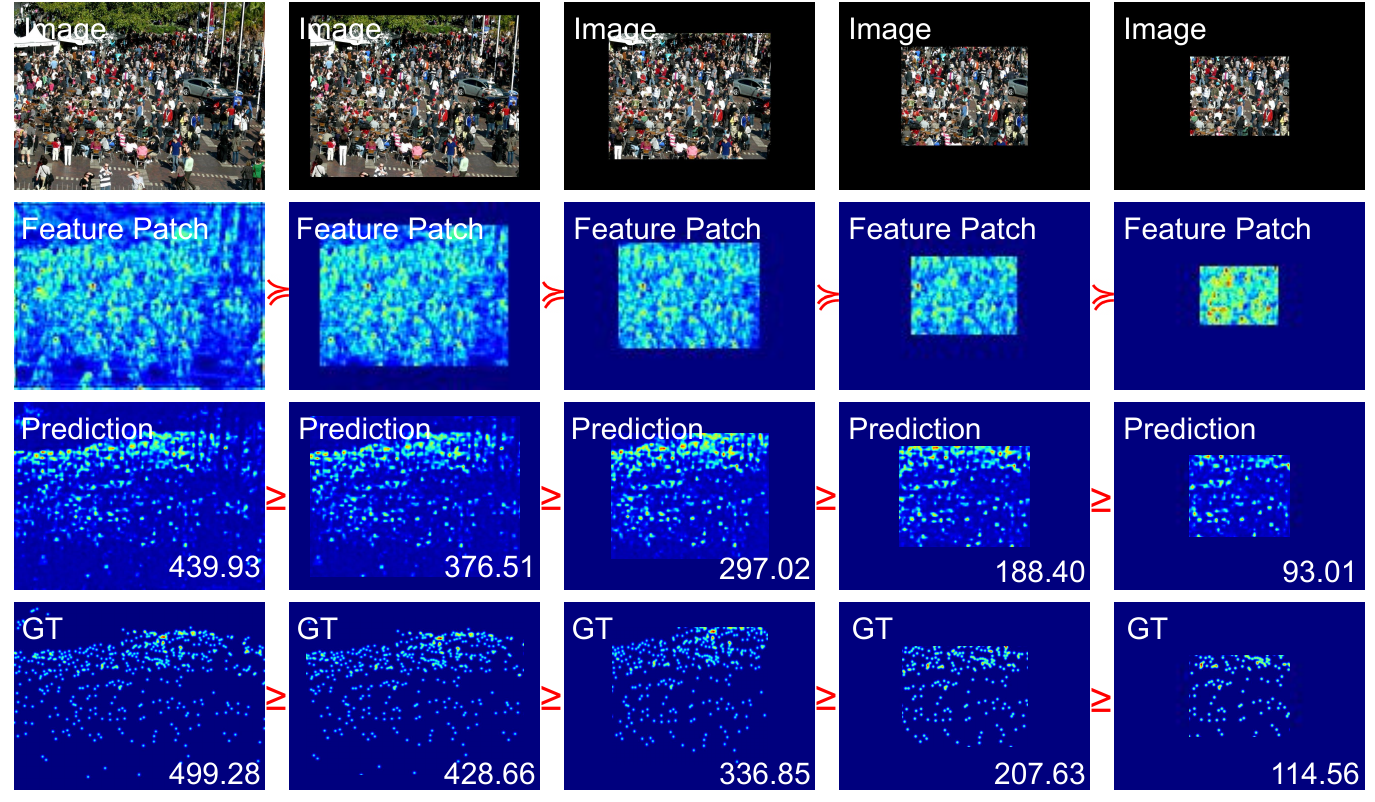}
   \caption{\gao{The illustration of potential partial orders existing at the feature level.} The images in the first row are the corresponding receptive fields of the cropped feature patches in the second row. Since the feature maps have 512 channels in practice, we just fuse these channels and visualize the mean feature map. The predictions and ground-truths of these cropped feature patches are shown in the last two rows. `$\succeq$' and `$\geq$' mean partial orders and fixed ordinal relations of these cropped feature patches, respectively.}
   \label{figures: motivation}
\end{figure}

\subsection{Motivation}\label{section: observation}
In the area of crowd counting, a common sense is that 
for any image of arbitrary size, the number of people in an image patch is always greater than or at least equal to the number in its sub-regions~\cite{liu2018leveraging}. 
Inspired by this observation and the definition of 
the receptive field, we believe this common sense is also valid at the feature level. This is because convolution and pooling layers in convolutional neural networks maintain the relative positions of the objects in an image. In other words, each position in the intermediate feature maps should represent the specific corresponding regions in the input space, which is often called the receptive field.
Thus, it should be guaranteed that the output counts predicted by larger feature patches in the latent space are no fewer than the counts predicted by their sub-patches in the same feature map, as shown in Fig.~\ref{figures: observation}.
Additionally, this consistency in partial orders is expected across different intermediate layers of the network.
In this way, we can greatly increase the utilization ratio of partial orders and structural information among unlabeled images. Further, we visualize the receptive field, feature patches in hidden layers, and the prediction and ground truth of our proposed method in Fig.~\ref{figures: motivation}. The visualization results clearly validate our hypothesis that partial orders persist in the feature layers.

\begin{figure*}[ht]
   \centering
   \includegraphics[width=0.95\linewidth]{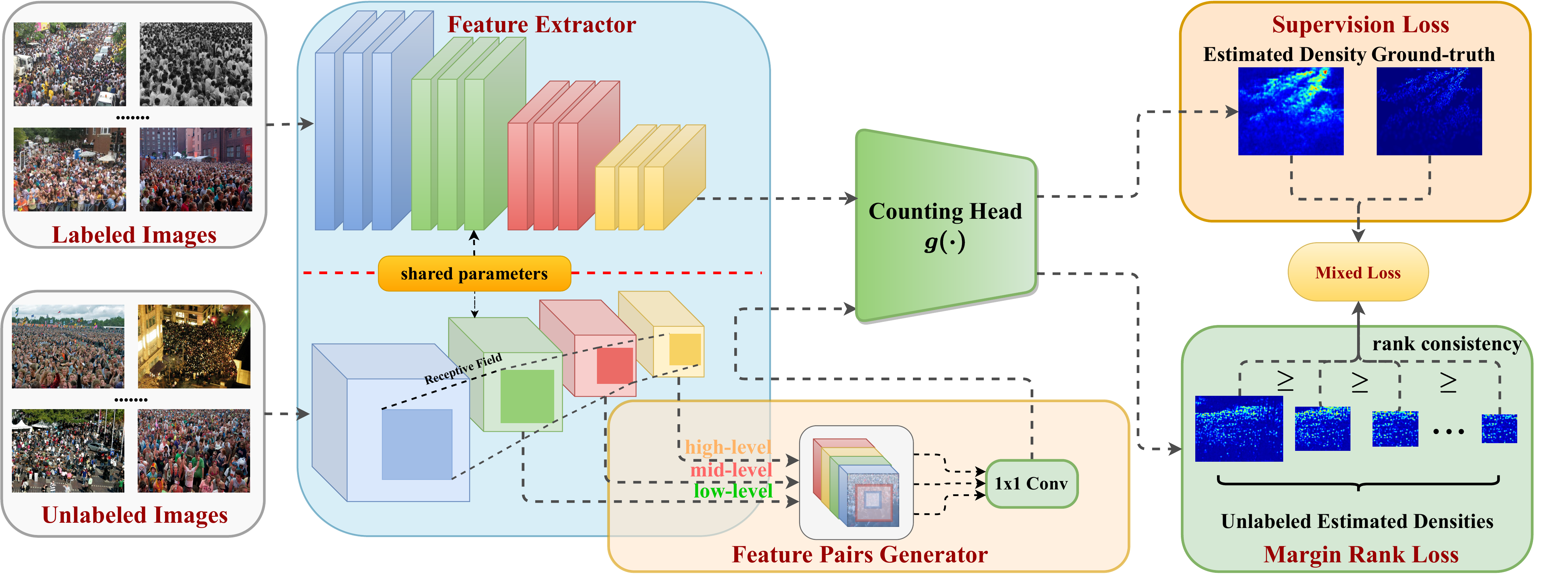}
   \caption{{\gao{The main architecture of our model.} The feature extractor module is shared for both labeled and unlabeled images to extract the latent features. The feature pairs generator is to prepare the feature patch pairs that satisfy the partial order relations. Mixed loss is made up of the fully-supervised counting loss (L2 loss) on labeled images and margin ranking loss on unlabeled images.}}
   \label{fig:model architecture}
\end{figure*}

\subsection{Feature Pairs Set Generation for Unlabeled Images}

To compute the margin rank loss of unlabeled images, a prerequisite is to obtain a set of feature patch pairs satisfying the \re{rank consistency}. We \re{generate the rank-consistent} feature pairs set from low-level to high-level layers. For any given feature map $F \in R^{C \times H \times W}$ within a given layer, we crop $M$ sub-regions, all sharing the same center point and with exponentially decreased cropped ratio $r^M$, where $r$ ranges between 0 and 1. The center point is randomly sampled from a small region centered in the feature map, {\re{whose size is $\frac{1}{8}H\times\frac{1}{8}W$.}}
After the cropping process, specifically, we guarantee that every sub-region is totally contained by its larger sub-region for training. Together with the feature map $F$ itself, we can obtain $M+1$ feature patches, with any two forming potential candidates for the rank-consistent feature pair set $S$, within a layer. Formally, we have
\begin{equation}
   S = \Big \{\langle v_m, v_n \rangle \big| m < n,  v_m \cap v_n = v_m, v_m \cup v_n = v_n\Big \}\;,
\end{equation}
where $m, n \in \{0,1,2,...,M\}$ and $M$ is set to 4 in practice. $v_i$ represents the cropped feature patches, with $v_0$ being the smallest feature patch and $v_M = F$ the largest. Therefore, each feature patch $v_m$ is wholly contained in feature patches $v_n$,  $\forall n>m$. 
More details of generating feature patch pairs set in one layer can be seen in Algorithm \textcolor{red}{1}.

\begin{table}[t]
\resizebox{\linewidth}{!}{
    \begin{tabular}{lp{180pt}}
       \toprule[1pt]
       \multicolumn{2}{l}{\textbf{Algorithm 1:} Generating Rank-Consistent Feature Patch Pairs Set}                                \\ \midrule[0.5pt]
       \textbf{Input :}  & A feature map $F \in R^{C \times H \times W}$, number of cropped patches $M$, cropped ratio $r$. \\
       \textbf{Step 1 :} &
         Choose a point as the center point in one small region randomly. The region is defined to be $r^M$ the shape of this feature map $F$, with the same aspect ratio, centered by point $(\frac{c}{2},\frac{h}{2},\frac{w}{2})$ . 
          \\
       \textbf{Step 2 :} & Initialize the rank-consistent feature pairs set $S = \varnothing$.                                    \\
       \textbf{Step 3 :} &  Choose the first and largest cropped patch $v_0 = F$.                                                 \\
       \textbf{Step 4 :} &
         Crop $M$ patches centered at the new center point. These $M$ patches $v_1, v_2, ... v_{M}$ are cropped by the ratio $r$ iteratively. \\
       \textbf{Step 5 :} & Resize these $M$ patches to the same size of $F$ and add each feature patch pair $\langle v_m, v_n\rangle$, $\forall m < n$ to the $S$. \\
       \textbf{Output :} & A set of feature patch pairs $S$.                                                         \\ \midrule[0.5pt]
       \textbf{Notations :}& $F \in R^{C \times H \times W}$ where $C, H, W$ represent the channels, height, and width of this feature map $F$ respectively. $(\frac{c}{2},\frac{h}{2},\frac{w}{2})$ is the coordinate of the center point. \\
       \bottomrule[1pt]
    \end{tabular}
}
\end{table}

\subsection{Model Architecture}

We describe the network architecture in detail here. Our network mainly consists
of two modules, the feature extractor and the crowd density map estimator. 
The feature extractor learns coarse-to-fine features through several convolutions
and max-pooling operations, while the crowd density map estimator regresses the 
density map based on these features. Our backbone of the feature extractor module is derived
from the VGG-16 network~\cite{simonyan2014vgg}. We only use the first ten layers of VGG-16
with pre-trained weights to train our feature extractor. 
\begin{equation}
   v^{(i)} = f(x^{(i)};\theta)\;,
\end{equation}
where the feature extractor module $f(\cdot;\theta)$ with the parameters $\theta$ contains the first ten layers of pretrained VGG-16 network. And the module learns the mapping function from the $i$-th input image $x^{(i)}$ to output its corresponding feature $v^{(i)}$.

The dilated convolutional layers
with $3\times 3$ kernel size, dilated ratio 2, and stride 1 followed by an upsampling layer
constitute the density map estimator, same as described by CSRNet-B~\cite{li2018csrnet}.
\begin{equation}
   D^{(i)} = g(v^{(i)};\phi)\;,
\end{equation}
where $D^{(i)}$ are the predicted
density map of the $i^{th}$ image. $g(\cdot;\phi)$ is the density estimator
with the parameters $\phi$ that contains six dilated convolutional 
layers, $1\times 1$ conv layer, and an upsampling layer.
We use the {\re{pixel-wise}} $L_2$ loss as our supervised loss, $L_s$, among labeled images. 
\begin{equation}
   L_s = \frac{1}{2N} \sum_{i=1}^N \bigg\|g\Big(f\big(x^{(i)};\theta\big);\phi\Big)- y^{(i)}\bigg\|^2_2\;,
\end{equation}
where $N$ is the number of training images in a batch, and $x^{(i)}$ and $y^{(i)}$ 
are the $i^{th}$ original input image and corresponding ground truth density map
in one batch, respectively. 

For training the unlabeled images, we use the same feature extractor architecture
with shared parameters shown in Fig.~\ref{fig:model architecture}. 

\subsection{Coarse-to-Fine Feature Pyramid Margin Ranking Loss}
In a deep CNN, feature maps are downsampled using convolutional or pooling layers. The receptive field shows that each pixel in the intermediate feature maps captures the information from one region of the input space. As discussed in Section~\ref{section: observation}, larger
regions contain the same number or more individuals compared to smaller sub-regions
within the input space. Similarly, because of the receptive field, the counts of 
the predicted density map from smaller regions of feature maps in the latent space 
should be equal to or less than those derived from super-regions of the same feature maps.

We adopt the margin ranking loss \re{for training the unlabeled images.}
We expect the network to learn the ordinal relations of \re{those cropped rank-consistent feature patch pairs} in the latent
space. Meanwhile, we should guarantee that \re{rank consistency} is maintained across 
pyramid features at different latent layers.
More specifically, we crop the feature maps among unlabeled images in $K$ latent layers and construct a margin ranking loss. For the $i^{th}$ unlabeled image, the loss $L_r^{(i)}$ is defined as follows:
\begin{equation}
   L_r^{(i)} = \max\Big(0, g\big(v^{(i)}_{u,m}\big) - g\big(v^{(i)}_{u,n}\big) + \epsilon\Big)\;,
\end{equation}
where $v^{(i)}_{u,m}$ and $v^{(i)}_{u,n}$ make up a rank-consistent feature patch pair of the $i^{th}$ unlabeled image from $S$ in the latent space. The subscript $u$ in $v_u$ denotes that the cropped patch is from an unlabeled image. The subscript $m$ and $n$ represent any two of the corresponding feature pairs. $\epsilon$ is the margin, indicating the model error-tolerance capacity. We expect the densities from a smaller region of feature maps $g(v^{(i)}_{u,m})$
in latent space to be no more than that from its super-region $g(v^{(i)}_{u,n})$.
Therefore, when the network predicts the correct ordinal relation $g(v^{(i)}_{u,m}) \leq g(v^{(i)}_{u,n})$,
the loss $L_r^{(i)}$ becomes zero, eliminating the need for gradient backpropagation. 
Otherwise, the loss $L_r^{(i)}$ captures the difference value between these two estimates, leading to gradient backpropagation that updates the model parameters. Consequently, the rank-consistent feature pyramid margin ranking loss among unlabeled images is defined as follows:
\begin{small}
\begin{equation}
\begin{aligned}
   L = \sum_{i=1}^N \sum_{k=1}^{K} \sum_{m=0}^{M-1} &\sum_{n=m+1}^{M} \max\Big(0, D_{u,m,k}^{(i)} - D_{u,n,k}^{(i)} + \epsilon\Big) \;,
\\ L_u &= \frac{2}{NKM(M+1)} L \;,
\end{aligned}
\end{equation}
\end{small}

\noindent where $N$ is the number of unlabeled images we used during training.
$K$ is the number of coarse-to-fine latent spaces selected to generate the rank-consistent feature pairs set $S$. $M$ is the number of cropped patches from the same latent space $k$. $D_{u,m,k}^{(i)}$ and $D_{u,n,k}^{(i)}$
are the density maps predicted by the ordinal relation feature pairs 
$\langle v^{(i)}_{u,m}, v^{(i)}_{u,n}\rangle$ in set $S$ in the $k^{th}$ latent space.

{\re{In our experiments, $M$ was set to 4 and $K$ to 3. Therefore, 
there are a total of $K$ feature pairs for a single image, and each pair contains $M+1$ feature regions~(the input feature maps are also included), leading to $K\times(M+1)$ predicted density maps in the margin ranking loss.
}}

We have introduced the rank-consistent feature pairs set generation method including the main architecture of our model, the fully-supervised loss $L_s$ for labeled images, and the rank-consistent margin ranking loss $L_u$ for unlabeled images. Thus, the final training loss we employed is the combination of $L_s$ and $L_u$, adjusted by the hyper-parameter $\lambda$. 
\begin{equation}
   L_{total} = L_s + \lambda L_u\;.
\end{equation}

\section{Datasets and Implementation Details}\label{section: implementation details}
\subsection{Experimental Setups.}
Existing semi-supervised crowd counting methods mainly follow two different training settings. First, off-the-shelf benchmark datasets--such as ShanghaiTech PartA~\cite{zhang2016single}, ShanghaiTech PartB~\cite{zhang2016single}, UCF-CC-50~\cite{idrees2013multi}, and UCF-QNRF~\cite{idrees2018composition}--are usually partitioned into labeled and unlabeled subsets with different proportions (namely, 5\%, 25\%, 30\%, and 50\% of the full dataset marked for labeled images). Second, other extra crowded images as the unlabeled data are collected for semi-supervised training. For a fair comparison, in this paper, we conducted experiments under both training settings, strictly following most previous literature. Unless noted otherwise, the 100\% labeled data under the semi-supervised mode has used our newly collected unlabeled dataset.

\subsection{Public Labeled Datasets}
\textbf{The UCF-CC-50 dataset~\cite{idrees2013multi}:} The UCF-CC-50 dataset is the first large-scale congested
scene dataset for pedestrian counting. Only 50 images with varying resolutions 
are collected among available images from public websites. An average of 1,279 persons 
appeared in each image, with counts ranging from a maximum of 4,543 to a minimum of 94. Both the small
number of images and the drastic variations in the number of people pose a considerable challenge for
the counting task. We used a 5-fold cross-validation approach for testing due to limited samples. We generated 
the density map as ground truth by geometry-adaptive Gaussian kernels~\cite{zhang2016single} for a fair comparison.

\textbf{ShanghaiTech PartA dataset~\cite{zhang2016single}:} The ShanghaiTech PartA dataset consists of 
482 images with varying resolutions. 241,677 heads are annotated in different illumination
conditions and crowded scenes. The density distributions span a broad spectrum, ranging from 
33 to 3,139 persons, with an average of 501 persons per image. We used geometry-adaptive 
kernels to generate the ground truth for all the images. For supervised learning,
researchers often split this dataset into two parts for training and testing. 

\textbf{ShanghaiTech PartB dataset~\cite{zhang2016single}:} The ShanghaiTech PartB dataset comprises
716 images with 88,488 head annotations captured from busy streets in the central business 
districts of Shanghai. The images have a fixed resolution of $768\times1024$ pixels. The ground truth
for these images is generated by a fixed Gaussian kernel with a variance $\sigma$ set at 15.

\textbf{The UCF-QNRF dataset~\cite{idrees2018composition}:} To construct a larger crowd counting dataset that includes a
dramatic variation of head sizes, diverse viewpoints and perspectives, and different locations and times of day, 1,535 images are collected from several search engines, including Google Image
Search and Flickr. Over 1,251,642 coordinates are labeled, costing more than 2,000
labor-hours. Due to the high resolution of the images, we limit the shorter side of a given image 
to a maximum of 1,920 pixels to fit the memory capacity. Images are rescaled with the same ratio to refrain from 
the global and local contextual information loss as little as possible. The training and test sets consist of 1,201
and 334 images, respectively. Similar to the ground truth generation method in the ShanghaiTech PartB 
dataset, a fixed Gaussian kernel is adopted.

\begin{figure}[h]
    \centering
    \includegraphics[width=\linewidth]{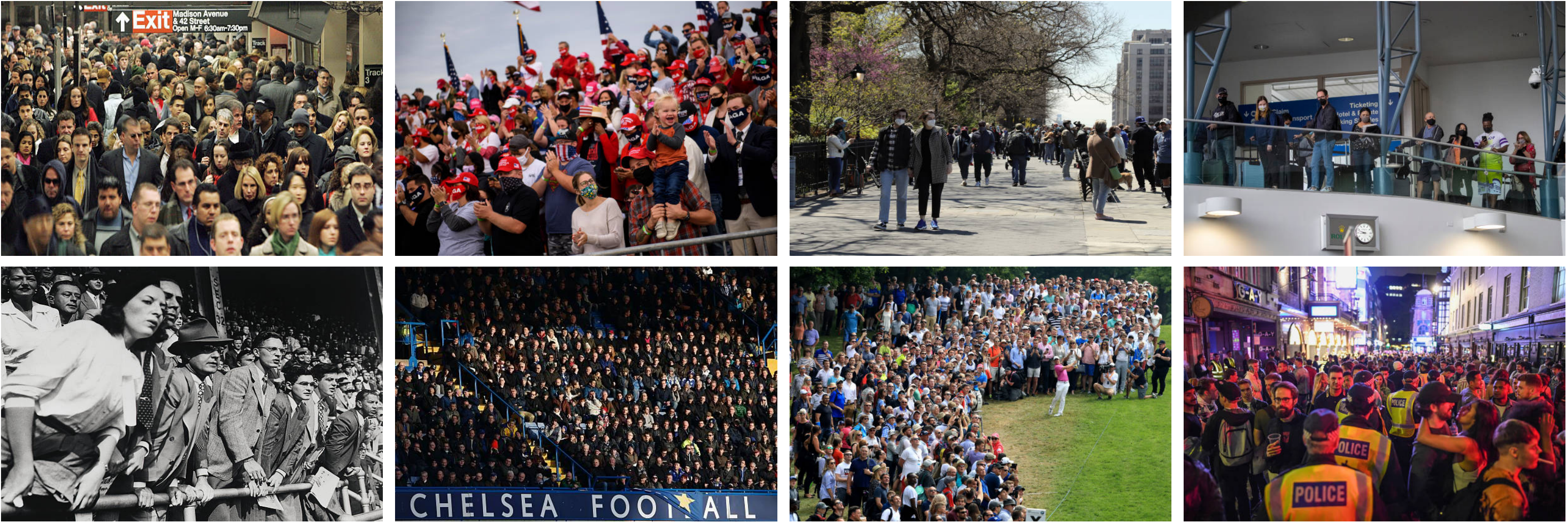}
    \caption{A few examples of our proposed FUDAN-UCC dataset captured from the images search engine GettyImages. It covers complex real-world scenarios, different people distribution, and various illumination conditions.}
    \label{fig:unlabeled}
\end{figure}

\subsection{FUDAN-UCC Unlabeled Dataset}
We captured 4,000 images in total from the GettyImages search engine using the keyword `crowd' to create the unlabeled dataset for our experiments. Fig.~\ref{fig:unlabeled} illustrates a few natural images of our collected dataset that contains varying scenarios, diverse people distribution, and different illuminations. Image resolutions range from $221\times 612$ to $612 \times 612$. This captured dataset may serve as a potential standard unlabeled crowd counting benchmark dataset for academic researchers to investigate semi-supervised or unsupervised learning methods in the future. \gao{The FUDAN-UCC dataset is publicly available,\footnote{URL: \url{https://1drv.ms/u/s!AkFyOI5q6fU_iGqJ6MibgUNja45x?e=crfMS0}} but is strictly intended for academic and non-commercial purposes. Anyone interested in using it should obtain the necessary licenses and comply with applicable laws and regulations before downloading.}

\subsection{Implementation Details}
We followed similar preprocessing methods in prior research~\cite{sindagieccv2020learning, yanliusemieccv2020} for supervised training. For semi-supervised learning, we randomly chose the training set with different proportions of labeled and unlabeled samples. Considering the images with larger resolutions of the UCF-QNRF dataset and memory constraints, we resized the shorter side to a maximum of 1,920 pixels and maintained the original aspect ratio. Effective data augmentations like random horizontal flipping, random cropping, and normalization were also adopted for both labeled and unlabeled images for training. The captured images were also resized with the same aspect ratio to fit the cropping operations. We used the Adam optimizer with the learning rate of $10^{-5}$ and weight decay of $10^{-4}$ in all of our experiments. {\re{For the rank-consistent feature pairs set generation process, we set $K$ to 3, \ie, low-level, mid-level, and high-level from the feature extractor, and the cropped ratio $r$ to 0.75.}}
\section{Experimental Results}
\label{section: results}

\subsection{Evaluation Metrics}
We employed two commonly-used evaluation metrics
(Mean Absolute Error, MAE, and Root Mean Squared Error, RMSE) to evaluate our model. 
The formulae of MAE and RMSE are defined as follows:
\begin{small}
\begin{equation}
   \text{MAE}=\frac{1}{N} \sum_{i=1}^N | \hat{Y_i} - Y_i |\;,
   \text{RMSE}=\sqrt{\frac{1}{N} \sum_{i=1}^N \big\| \hat{Y_i} - Y_i \big\|^2_2}\;,
\end{equation}
\end{small}
where $N$ denotes the number of images from the test set. $\hat{Y_i}$ and $Y_i$
are the predicted counts and actual counts of the $i$-th image, respectively.
Briefly speaking, MAE implies the precision of the estimates, and RMSE implies
the robustness of the estimates. RMSE is more sensitive to the outliers.
Obtaining a model with a low MAE as well as a low RMSE is our expectation.

\subsection{Evaluation on the UCF-CC-50 dataset:}
The experimental results on the UCF-CC-50 dataset are shown in Table~\ref{results: ucfcc}.
Since there are only 50 images in the UCF-CC-50 dataset, it is not suitable for dividing
into labeled and unlabeled datasets. We just utilized the unlabeled images from the collected dataset to train our model. For a fair comparison, we used the same
5-fold cross-validation to compute the average MAE and average RMSE metrics. The baseline model
was trained by using only labeled 50 images in a fully-supervised way. It achieved 266.10 average MAE. 
We reproduced the L2R method using our model architecture. It improved the performance on this dataset and
led to 261.60 average MAE. Our approach thus achieved an MAE improvement of nearly 10 compared with the L2R method.

\begin{table}[ht]
   \caption{Estimation errors on the UCF-CC-50 dataset. The baseline model is trained on only 50 images following a fully-supervised framework. For the `semi', we treat the same 50 images on the UCF-CC-50 dataset as labeled images, and leverage unlabeled images from our collected dataset. A 5-fold cross-validation approach is used among them.}
   \centering
   \begin{tabular}{lccc}
   \toprule[1pt]
   \multirow{2}{*}{Method} & \multicolumn{3}{c}{UCF-CC-50}  \\ \cmidrule{2-4}
                      & Mode           & \textit{Avg.} MAE             & \textit{Avg.} RMSE            \\ \midrule[0.5pt]
   Baseline  & Fully                   & 266.10                & 397.50                 \\
   L2R~\cite{liu2019exploiting} & Semi                    & 261.60                & 368.07                \\
   \methodname~(Ours)    & Semi                           & \textbf{251.52}                & \textbf{341.06}                \\
   \bottomrule[1pt]                 
   \end{tabular}
   \label{results: ucfcc}
\end{table}

\begin{table*}[ht]
\centering
\caption{Estimation errors on the ShanghaiTech PartA, ShanghaiTech PartB and UCF-QNRF datasets. L2R$^*$ is our reproduced results.}
\label{results: all}
\begin{tabular}{lcccccccc}
\toprule[1pt]
\multirow{2}{*}{Method}                                                           & \multirow{2}{*}{Labeled images} & \multirow{2}{*}{Mode} & \multicolumn{2}{c}{ShanghaiTech PartA} & \multicolumn{2}{c}{ShanghaiTech PartB} & \multicolumn{2}{c}{UCF-QNRF}    \\\cmidrule{4-9} & & & MAE                & RMSE              & MAE                & RMSE              & MAE            & RMSE           \\\midrule[0.5pt]
Baseline                                                                          & \multirow{4}{*}{5\%}            & Fully                 & 118.0              & 211.0             & 21.2               & 34.2              & 186.0          & 295.0          \\
L2R~\cite{liu2019exploiting}                                &                                 & Semi                  & 115.0              & 208.0             & 20.1               & 32.9              & 182.0          & 291.0          \\
Sindagi \etal~\cite{sindagieccv2020learning} &                                 & Semi                  & \textbf{102.0}     & \underline{172.0}       & \textbf{15.7}      & \textbf{27.9}     & \textbf{160.0} & \textbf{275.0} \\
\methodname~(Ours)                                            &                                 & Semi                  & \underline{112.7}        & \textbf{165.7}    & \underline{19.8}         & \underline{30.3}        & \underline{173.4}    & \underline{279.8}    \\\midrule[0.5pt]
Baseline                                                                          & \multirow{3}{*}{25\%}           & Fully                 & 110.0              & 160.0             & -                  & -                 & 178.0          & 252.0          \\
Sindagi \etal~\cite{sindagieccv2020learning} &                                 & Semi                  & \textbf{91.0}      & \underline{149.0}       & -                  & -                 & \underline{147.0}    & \textbf{226.0} \\
\methodname~(Ours)                                            &                                 & Semi                  & \underline{93.5}         & \textbf{148.4}    & -                  & -                 & \textbf{146.9} & \underline{237.2}    \\\midrule[0.5pt]
Baseline                                                                          & \multirow{4}{*}{30\%}           & Fully                 & 98.3               & 159.2             & 15.8               & 25.0              & 147.7          & 253.1          \\
L2R~\cite{liu2019exploiting}                                &                                 & Semi                  & 90.3               & 153.5             & 15.6               & 24.4              & 148.9          & 249.8          \\
IRAST~\cite{yanliusemieccv2020}                             &                                 & Semi                  & \underline{86.9}         & \underline{148.9}       & \textbf{14.7}      & \textbf{22.9}     & \textbf{135.6} & \underline{233.4}    \\
\methodname~(Ours)                                            &                                 & Semi                  & \textbf{86.5}      & \textbf{121.2}    & \underline{15.1}         & \underline{23.8}        & \underline{137.4}    & \textbf{230.0} \\\midrule[0.5pt]
Baseline                                                                          & \multirow{4}{*}{50\%}           & Fully                 & 102.0              & 149.0             & 15.9               & 25.7              & 158.0          & 250.0          \\
Sindagi \etal~\cite{sindagieccv2020learning} &                                 & Semi                  & 89.0               & 148.0             & 16.8               & 25.1              & 136.0          & \underline{218.0}    \\
SUA~\cite{meng2021spatial}                                  &                                 & Semi                  & \textbf{68.5}      & \underline{121.9}       & \underline{14.1}         & \underline{20.6}        & \underline{130.3}    & 226.3          \\
\methodname~(Ours)                                            &                                 & Semi                  & \underline{78.4}         & \textbf{112.9}    & \textbf{9.8}       & \textbf{15.3}     & \textbf{130.2} & \textbf{203.8} \\\midrule[0.5pt]
Baseline                                                                          & \multirow{6}{*}{100\%}          & Fully                 & 69.1               & 103.0             & 10.6               & 16.0              & 119.2          & 211.4          \\
SUA~\cite{meng2021spatial}                                  &                                 & Fully                 & 66.9               & 125.6             & 12.3               & 17.9              & 119.2          & 213.3          \\
STC-Crowd~\cite{wang2023semi}                               &                                 & Fully                 & \textbf{62.3}      & \underline{103.5}       & 9.5                & 14.6              & 97.6           & \underline{160.3}    \\
L2R~\cite{liu2019exploiting}                                &                                 & Semi                  & 73.6               & 106.6             & 13.7               & 21.4              & -              & -              \\
L2R$^*$~\cite{liu2019exploiting}                            &                                 & Semi                  & 64.4               & 105.0             & \underline{8.2}          & \underline{13.9}        & \underline{97.1}     & 169.9          \\
\methodname~(Ours)                                            &                                 & Semi                  & \textbf{62.6}      & \textbf{102.0}    & \textbf{7.9}       & \textbf{13.4}     & \textbf{94.0}  & \textbf{159.8}\\
\bottomrule[1pt]
\end{tabular}
\end{table*}

\begin{table*}[ht]
\centering
\begin{minipage}[c]{0.35\linewidth}
\caption{Different utilization ratios of labeled images on the ShanghaiTech PartA dataset.}
\resizebox{\linewidth}{!}{
\begin{tabular}{crrrr}
   \toprule[1pt]
        \multirow{2}{*}{utilization ratio}& \multicolumn{2}{c}{label-only} & \multicolumn{2}{c}{label+unlabel}\\\cmidrule{2-5}
        &MAE &RMSE    & MAE & RMSE   \\\midrule[0.5pt]
       5\% &118.0 &211.0 &112.7  &165.7    \\
       25\%&110.0 &160.0 & 93.5  &148.4    \\
       50\%&102.0 &149.0 & 78.4    & 112.9    \\\midrule[0.5pt]
       100\%     &69.1  &103.0 &62.6 &102.0   \\
   \bottomrule[1pt]
\end{tabular}
}
\label{results: different ratio}
\end{minipage}
\begin{minipage}[c]{0.25\linewidth}
\caption{Impact of $\lambda$ in the loss on the ShanghaiTech PartA dataset.}
\resizebox{\linewidth}{!}{
\begin{footnotesize}
\begin{tabular}{ccccc}
   \toprule[1pt]
      & $\lambda$ & MAE & RMSE &  \\\midrule[0.5pt]
      & 0.1     & 69.3    & 104.2     &  \\
      & 0.5     & 67.4    & 112.1     &  \\
      & 1     & \textbf{62.6}    & \textbf{102.0}     &  \\
      & 5     & 63.7   & 105.4     &  \\
      & 10    & 65.2    & 106.3     &  \\
   \bottomrule[1pt]
\end{tabular}
\end{footnotesize}
}
\label{results: lambda}
\end{minipage}
\begin{minipage}[c]{0.3\linewidth}
\caption{Combination of rank losses in different layers on the ShanghaiTech PartA dataset. }
\centering
\begin{tabular}{ccccc}
   \toprule[1pt]
      low & mid & high & MAE & RMSE \\\midrule[0.5pt]
\checkmark&    &    &    \gao{65.5}&\gao{103.8}     \\
&\checkmark    &    &    \gao{64.1}&\gao{103.1}     \\

             &                    & \checkmark           &63.6     & 108.4     \\
             & \checkmark                & \checkmark           & 63.4    & 107.0     \\
      \checkmark       & \checkmark                & \checkmark        & \textbf{62.6}   & \textbf{102.0}     \\
   \bottomrule[1pt]  
\end{tabular}
\label{results: combination}
\end{minipage}
\end{table*}

\subsection{Evaluation on the ShanghaiTech dataset:}
We demonstrate the experimental results on both the ShanghaiTech PartA and PartB datasets 
in Table~\ref{results: all}, respectively.
All labeled and unlabeled images were chosen from the ShanghaiTech datasets. We randomly
picked 5\%, 25\%, and 50\% labeled images from the training set as labeled samples, while the rest of the training set
were regarded as unlabeled samples for training.
We compared our method against five previous methods, L2R~\cite{liu2019exploiting}, Sindagi~\etal~\cite{sindagieccv2020learning},
IRAST~\cite{yanliusemieccv2020}, and SUA~\cite{meng2021spatial}, {\re{and STC-Crowd~\cite{wang2023semi}}}. These methods used
different ratios of labeled images to train their models and reported their results.
For a fair comparison to them, we used the same proportion of labeled images.
For PartA, we reached the best results when the ratio was 50\% with MAE 78.4. For 5\% and 25\%
settings, we obtained competitive results for MAE and RMSE compared with Sindagi \etal~\cite{sindagieccv2020learning}. A possible explanation is that 5\% means only 15 labeled images for training, and the ranking loss among the remaining unlabeled images only reflects qualitative relations which cannot help the model predict the specific count number accurately with a small number of labeled images. As for the PartB dataset, we observed that partial orders among unlabeled images are of limited benefit, especially when the number of labeled samples in the training dataset was relatively small. This can be attributed to two primary reasons: 1) the density distribution is relatively sparse in this dataset and the number of people is relatively small, and 2) our proposed qualitatively partial orders among unlabeled images may be more suitable and efficient for crowded scenes. Nevertheless, our proposed method still achieved performance comparable to other methods. When we used all images from the training set and our collected unlabeled ones for training, the results of \methodname~achieved the lowest MAE and RMSE scores.

Our model achieved the most significant performance improvements with 50\% labeled images, compared to 5\% or 100\%.
It is worth emphasizing that although the proposed method does not require human annotations for unlabeled data, it still benefits from the increasing number of labeled training data. 
Using only 5\% labeled data does not provide the model with sufficient supervision signals. On the contrary, increasing the labeled data to 50\%~(10 times than 5\%) significantly improves the performance. The performance gain decreases when further increasing the labeled data to 100\%~(only 2 times for 50\% to 100\%).

\begin{figure*}[ht]
   \centering
   \includegraphics[width=0.98\linewidth]{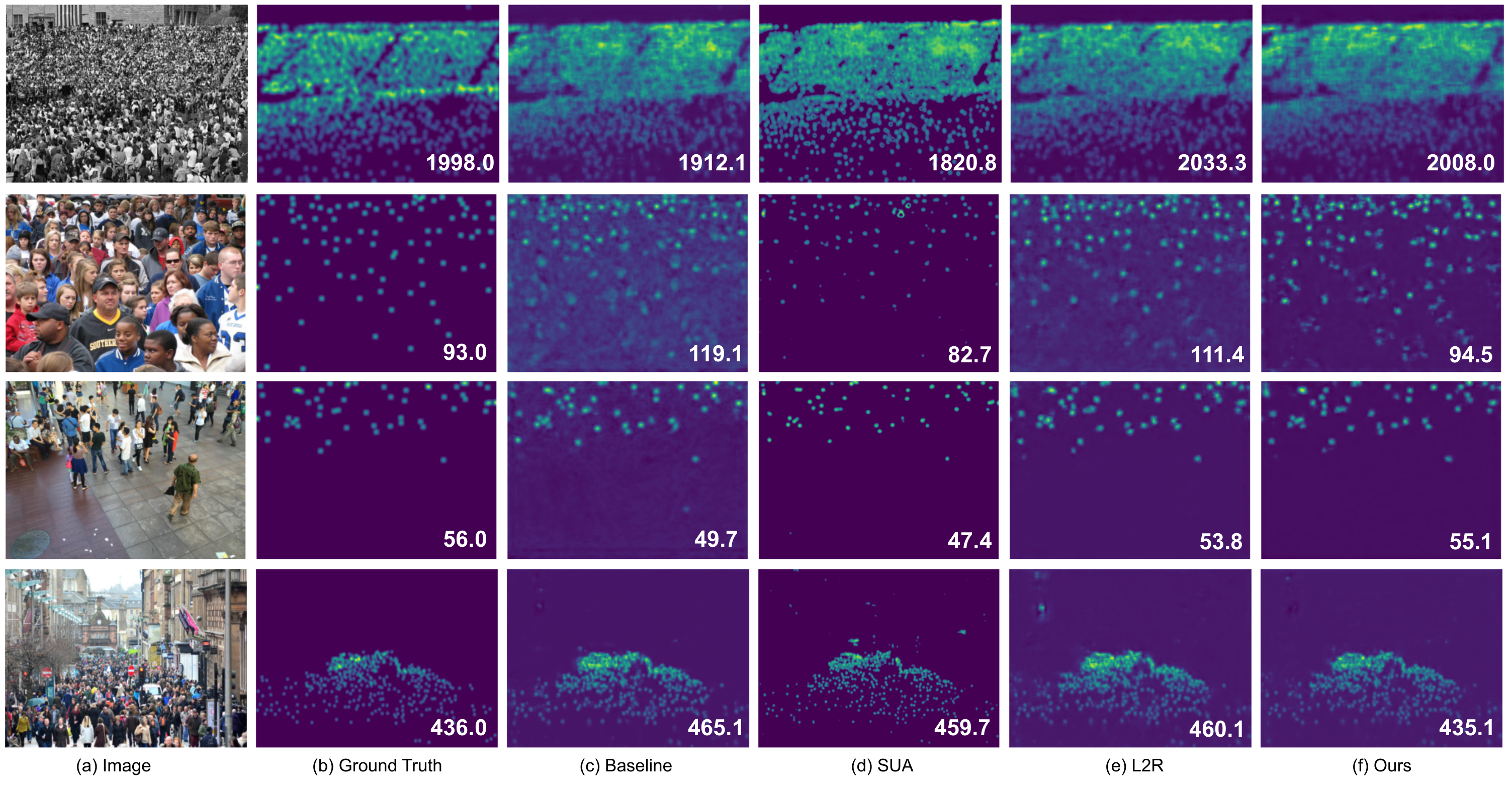}
   \caption{
   {\re{
   The visual comparisons of different competitors including the baseline, SUA~\cite{meng2021spatial}, L2R~\cite{liu2019exploiting}, and our method. The four images come from UCF-CC-50, ShanghaiTech PartA, UCF-QNRF, and ShanghaiTech PartB datasets, respectively.
   }}
   }
   \label{figures: visualization}
\end{figure*}
\subsection{Evaluation on the UCF-QNRF dataset:} The experimental results on the UCF-QNRF dataset are shown in Table~\ref{results: all}. We also randomly chose 5\%, 25\%, 30\%, and 50\% labeled images of its training set and regarded the rest as unlabeled ones for semi-supervised training to compare with previous methods~\cite{liu2018leveraging, sindagieccv2020learning, yanliusemieccv2020}. Moreover, we reproduced the L2R~\cite{liu2019exploiting} method using our collected unlabeled dataset. The results indicate that our proposed \methodname~is superior to the previous semi-supervised methods.

\subsection{Ablation Study}
\textbf{Different utilization ratios of labeled images: }
We designed the ablation study to verify whether our approach would be robust under the different settings
of the varying number of labeled images. We conducted this experiment on the ShanghaiTech PartA dataset.
We randomly chose 5\%, 25\%, 50\% images, respectively, to make up the labeled dataset and the rest images from the training set were the unlabeled ones. The unlabeled images from FUDAN-UCC dataset were only used when the utilization ratio was 100\%. As shown in Table~\ref{results: different ratio}, our method achieved a consistent
performance improvement towards training with only labeled images.

\textbf{Impact of varying $\lambda$:} Further, to exploit the role of margin ranking loss in the final mixed loss,
we tried different hyper-parameters $\lambda$ which represents the weight of self-supervised loss.
The value of $\lambda$ was chosen from $\{0.1, 0.5, 1, 5, 10\}$. The specific performance with different values of $\lambda$ is shown in Table~\ref{results: lambda}. We used the unlabeled images from our collected dataset together with all images from the training set of the ShanghaiTech PartA dataset for evaluating the impact of $\lambda$. Our model achieved the best performance on the ShanghaiTech PartA dataset when $\lambda$ is set to 1.

\begin{figure}[h]
   \centering
   \includegraphics[width=\linewidth]{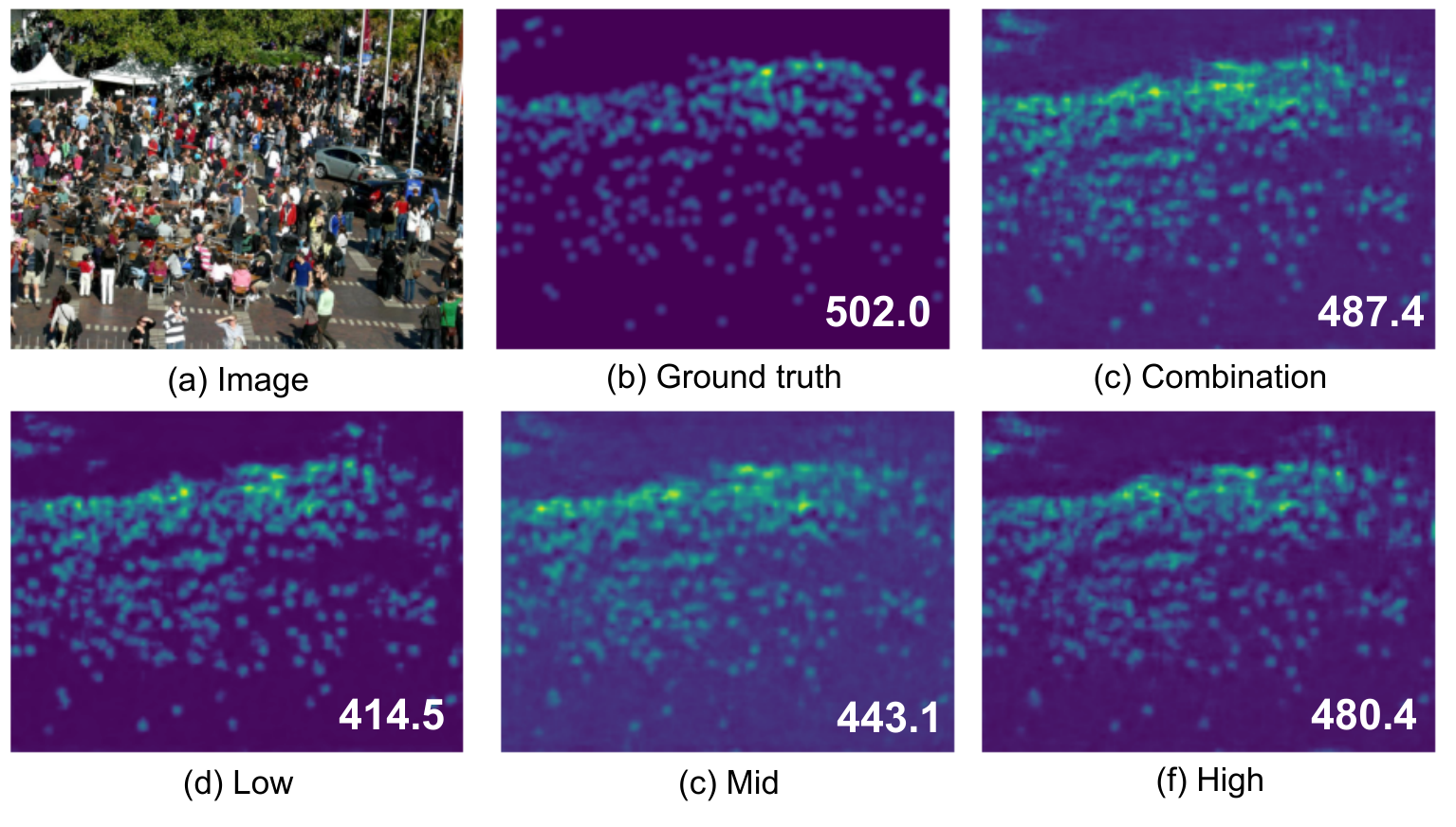}
   \caption{
   The visualization results of our proposed rank-consistent feature ranking loss applied to different feature levels.
   }
   \label{figures: abvis}
\end{figure}

\textbf{Combination of ranking losses in different layers: }
Coarse-to-fine pyramid features in different layers represent high-level semantic knowledge and low-level visual signals including textures, edges, backgrounds, among others. Moreover, partial orders should exist in feature patches from different layers. Therefore, we report the different results caused by a diverse combination of ranking loss in disparate layers (low-level, mid-level, and high-level), as shown in Table~\ref{results: combination}. 
We discovered that the performance would be better as the utilization ratio of coarse-to-fine pyramid feature patches with partial orders increases. In Fig.~\ref{figures: abvis}, we also visualize the results of the ranking loss used in different layers. We observed that the low level with the worst performance could not capture the semantic information for crowd counting, the middle level produced a uniform density map, and the high level might be the most helpful than the other two feature levels. Combining the three feature levels achieved the best counting performance with high-quality estimated density maps.

\textbf{Different unlabeled datasets:} We conducted an ablation study to show the effectiveness of our proposed FUDAN-UCC as an unlabeled dataset. Specifically, under the same setting in our paper, the ShanghaiTech PartA dataset was adopted as the labeled dataset while the ShanghaiTech PartB, UCF-QNRF, and FUDAN-UCC were employed as unlabeled datasets for fair comparisons, respectively.
As shown in Table~\ref{table:abltion_unlabeled_data},
compared to the baseline, our proposed FUDAN-UCC made the greatest improvements over other datasets, showing its effectiveness for semi-supervised crowd counting.

\begin{table}[h]
\centering
\caption{The ablation study of different unlabeled datasets. 
Baseline denotes that no additional unlabeled datasets are used. 
}
\label{table:abltion_unlabeled_data}
\begin{tabular}{lcc}
\toprule[1pt]             
Unlabeled dataset    & MAE  & RMSE  \\
\midrule
Baseline & 69.1 & 103.0 \\
ShanghaiTech PartB & 65.3 & 117.7 \\
UCF-QNRF           & 68.9 & 108.8      \\
FUDAN-UCC          & \textbf{62.6} & \textbf{102.0} \\
\bottomrule[1pt]             
\end{tabular}

\end{table}

\textbf{Margin ranking loss on different training data:} We conducted the experiments when adding this margin loss to the different combinations of training data in Table~\ref{table:ablation_ranking1}.
Compared to the baseline, the results suggest that the performance can be improved by introducing margin ranking loss. Leveraging more unlabeled images can further boost the performance compared to train on only labeled images.
Applying the margin ranking loss to both unlabeled and labeled data does not exhibit significant differences from the ones on only unlabeled data. We argue that it is because ranking consistency between different feature pairs can be easily learned by the counting model when the supervision training signals exist. Consequently, the margin ranking loss hardly contributes to the counting model using the labeled data.

\begin{table}[h]
\centering
\caption{
The ablation study of the margin ranking loss applied to different combinations of training data.
}
\resizebox{\linewidth}{!}{
\begin{tabular}{lcccccc}
\toprule[1pt]
& \multicolumn{2}{c}{ShanghaiTech PartA} & \multicolumn{2}{c}{ShanghaiTech PartB} & \multicolumn{2}{c}{UCF-QNRF} \\
\cmidrule{2-7}
& MAE               & RMSE               & MAE                & RMSE              & MAE           & RMSE         \\
\cmidrule{1-7}
Baseline & 69.1              & 103.0              & 10.6               & 16.0              & 119.2         & 211.4        \\
Labeled only & 66.2              & 118.9              & 10.0               & 15.2              & 98.8          & 173.1        \\
Unlabeled+labeled  & 63.7              & 106.4              & 8.1                & \textbf{12.5}              & 94.7          & 163.4      \\  
Unlabeled only~(Ours)  & \textbf{62.6}  & \textbf{102.0}              & \textbf{7.9}                & 13.4              & \textbf{94.0}          & \textbf{159.8}        \\
\bottomrule[1pt]
\end{tabular}
}
\label{table:ablation_ranking1}
\end{table}

\subsection{Visualization and Analysis}
We visualize the results on these four different datasets in Fig.~\ref{figures: visualization}. The visualization results clearly demonstrate the effectiveness of our proposed \methodname~model. The performance of utilizing the rank consistency among unlabeled images is better than that of training on labeled ones only~(Baseline). To be more specific, in the first and second row of Fig.~\ref{figures: visualization}, the baseline predicted density map is prone to learn the uniform distribution of crowded people especially in the crowded regions, training on labeled images only. {\re{The rest three methods can address this problem by adding more unlabeled data. However, our proposed method can achieve the best performance with better estimated density maps by utilizing the rank consistency at pyramid features.}}

\subsection{Limitations and Discussions}
In this paper, we introduce a deep rank-consistent pyramid model called \methodname~for improving crowd counting especially when the labeled images are limited. According to the analysis of extensive experimental results and visualization comparisons, \methodname~can rival state-of-the-art methods. However, it has certain limitations, which we will discuss in this section.
\subsubsection{{Failure cases}} 
\begin{figure}[h]
    \centering
    \includegraphics[width=\linewidth]{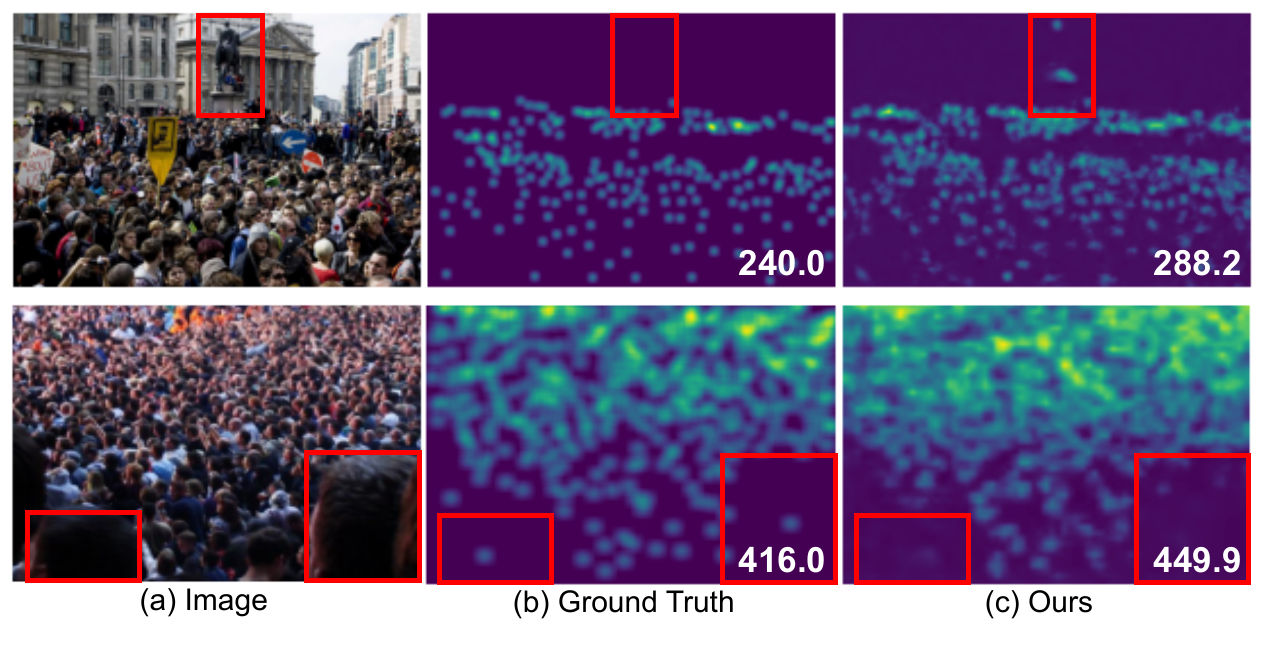}
    \caption{
    {\re{
    Failure cases visualization. Our model cannot distinguish the sculpture~(first row) and detect back-side heads~(second row).
    }}
    }
    \label{fig:limitations}
\end{figure}

Although \methodname~achieves superior estimated counts compared to baseline and other methods~(See Fig.~\ref{figures: visualization}), we have identified instances where it faltered in Fig.~\ref{fig:limitations}: 1) our model cannot distinguish real people and human-like sculpture; 
and 2) our model cannot detect patchy back-side heads, which appear almost completely black.
\subsubsection{{Imperfect annotations}} Due to the costly pixel-wise labeling process, imperfect annotations are bound to occur in public datasets. These include missing or ambiguous annotations, background noises, annotation shifts, and the like. In this paper, we mainly focus on the semi-supervised setting for crowd counting without considering the issue with noisy annotations. 
However, we believe that our method can mitigate such issues to a degree because the proposed rank-consistent model does not use manual annotations among unlabeled images during training.
In addition, our model is orthogonal to the current literature. We can integrate ideas from other techniques such as modeling noises~\cite{wan2020modeling} and cross-head~\cite{dai2023cross} to address these challenges.

\subsubsection{{FUDAN-UCC vs. Synthetic Datasets}} In this study, we collected an unlabeled dataset, the FUDAN-UCC, to assist the crowd counting model training when the labeled data is limited. The data-hungry issue of crowd counting remains challenging because of the time-consuming labeling process, especially for densely populated scenes. To address this issue, some researchers~\cite{wang2019learning, lian2021locating} introduced a few virtual datasets with synthetic images to provide sufficient training data. Wang~\etal~\cite{wang2019learning}~generated a GCC dataset with different people distribution, backgrounds, weather conditions, illumination, and camera perspectives by changing the hyperparameters of the game environment GTA5. Lian~\etal~\cite{lian2021locating}~provided the ShanghaiTechRGBD-Syn dataset with depth prior information and designed a depth-adaptive kernel to generate high-fidelity density maps for better training.
Unfortunately, the significant domain gap between synthetic and real-world data is inevitably introduced, which makes it challenging to simply train on synthetic data for better performance.
Instead of synthesizing training data, our motivation to collect FUDAN-UCC is that, as a real-world dataset, FUDAN-UCC could be directly used for assisting the model training without making any domain transfer. We believe that it could serve as a public real-world unlabeled dataset for comparing future semi-supervised crowd counting methods.

\section{Conclusion}
\label{section: conclusion}

Our work focused on taking advantage of partial orders from coarse-to-fine pyramid features to assist the neural network to enhance the qualitative discrimination among unlabeled images. Extensive experiments show that \methodname~outperforms other state-of-the-art methods with the help of self-supervised coarse-to-fine feature pyramid ranking loss, especially in dense crowd scenes.  Being simple and intuitive, our proposed method is easy to implement. Besides, our new dataset FUDAN-UCC can be a valuable addition to the community that could redefine the benchmark for unlabeled dataset in semi-supervised crowd counting. We believe the insights gained and potential solutions discussed provide a foundation for future innovations in this space.

\ifCLASSOPTIONcaptionsoff
  \newpage
\fi

\bibliographystyle{IEEEtran}
\bibliography{main}

\begin{thebibliography}{10}
\providecommand{\url}[1]{#1}
\csname url@samestyle\endcsname
\providecommand{\newblock}{\relax}
\providecommand{\bibinfo}[2]{#2}
\providecommand{\BIBentrySTDinterwordspacing}{\spaceskip=0pt\relax}
\providecommand{\BIBentryALTinterwordstretchfactor}{4}
\providecommand{\BIBentryALTinterwordspacing}{\spaceskip=\fontdimen2\font plus
\BIBentryALTinterwordstretchfactor\fontdimen3\font minus
  \fontdimen4\font\relax}
\providecommand{\BIBforeignlanguage}[2]{{%
\expandafter\ifx\csname l@#1\endcsname\relax
\typeout{** WARNING: IEEEtran.bst: No hyphenation pattern has been}%
\typeout{** loaded for the language `#1'. Using the pattern for}%
\typeout{** the default language instead.}%
\else
\language=\csname l@#1\endcsname
\fi
#2}}
\providecommand{\BIBdecl}{\relax}
\BIBdecl

\bibitem{leibe2005pedestrian}
B.~Leibe, E.~Seemann, and B.~Schiele, ``{Pedestrian Detection in Crowded
  Scenes},'' in \emph{Proceedings of the IEEE Conference on Computer Vision and
  Pattern Recognition}, vol.~1, 2005, pp. 878--885.

\bibitem{dollar2011pedestrian}
P.~Dollar, C.~Wojek, B.~Schiele, and P.~Perona, ``{Pedestrian Detection: An
  Evaluation of the State of the Art},'' \emph{IEEE Transactions on Pattern
  Analysis and Machine Intelligence}, vol.~34, no.~4, pp. 743--761, 2011.

\bibitem{chan2009bayesian}
A.~B. Chan and N.~Vasconcelos, ``{Bayesian Poisson Regression for Crowd
  Counting},'' in \emph{Proceedings of the IEEE International Conference on
  Computer Vision}, 2009, pp. 545--551.

\bibitem{ryan2009crowd}
D.~Ryan, S.~Denman, C.~Fookes, and S.~Sridharan, ``{Crowd Counting using
  Multiple Local Features},'' in \emph{Digital Image Computing: Techniques and
  Applications}, 2009, pp. 81--88.

\bibitem{chen2012feature}
K.~Chen, C.~C. Loy, S.~Gong, and T.~Xiang, ``{Feature Mining for Localised
  Crowd Counting},'' in \emph{Proceedings of The British Machine Vision
  Conference}, vol.~1, no.~2, 2012.

\bibitem{ren2015faster}
S.~Ren, K.~He, R.~Girshick, and J.~Sun, ``{Faster R-CNN: Towards Real-Time
  Object Detection with Region Proposal Networks},'' in \emph{Advances in
  Neural Information Processing Systems}, 2015, pp. 91--99.

\bibitem{redmon2016you}
J.~Redmon, S.~Divvala, R.~Girshick, and A.~Farhadi, ``{You Only Look Once:
  Unified, Real-Time Object Detection},'' in \emph{Proceedings of the IEEE
  Conference on Computer Vision and Pattern Recognition}, 2016, pp. 779--788.

\bibitem{redmon2018yolov3}
J.~Redmon and A.~Farhadi, ``{YOLOv3: An Incremental Improvement},'' \emph{arXiv
  preprint arXiv:1804.02767}, 2018.

\bibitem{liu2016ssd}
W.~Liu, D.~Anguelov, D.~Erhan, C.~Szegedy, S.~Reed, C.-Y. Fu, and A.~C. Berg,
  ``{SSD: Single Shot Multibox Detector},'' in \emph{Proceedings of The
  European Conference on Computer Vision}, 2016, pp. 21--37.

\bibitem{jiao2021new}
L.~Jiao, R.~Zhang, F.~Liu, S.~Yang, B.~Hou, L.~Li, and X.~Tang, ``{New
  Generation Deep Learning for Video Object Detection: A Survey},'' \emph{IEEE
  Transactions on Neural Networks and Learning Systems}, vol.~33, no.~8, pp.
  3195--3215, 2022.

\bibitem{cao2021hierarchical}
J.~Cao, Y.~Pang, J.~Han, and X.~Li, ``{Hierarchical Regression and
  Classification for Accurate Object Detection},'' \emph{IEEE Transactions on
  Neural Networks and Learning Systems}, vol.~34, no.~5, pp. 2425--2439, 2022.

\bibitem{wu2022enhanced}
Z.~Wu, J.~Wen, Y.~Xu, J.~Yang, X.~Li, and D.~Zhang, ``{Enhanced Spatial Feature
  Learning for Weakly Supervised Object Detection},'' \emph{IEEE Transactions
  on Neural Networks and Learning Systems}, 2022.

\bibitem{li2022dagcn}
C.~Li, F.~Liu, Z.~Tian, S.~Du, and Y.~Wu, ``{DAGCN: Dynamic and Adaptive Graph
  Convolutional Network for Salient Object Detection},'' \emph{IEEE
  Transactions on Neural Networks and Learning Systems}, 2022.

\bibitem{tan2022mhsa}
H.~Tan, X.~Liu, B.~Yin, and X.~Li, ``{MHSA-Net: Multihead Self-Attention
  Network for Occluded Person Re-Identification},'' \emph{IEEE Transactions on
  Neural Networks and Learning Systems}, 2022.

\bibitem{miao2021identifying}
J.~Miao, Y.~Wu, and Y.~Yang, ``{Identifying Visible Parts via Pose Estimation
  for Occluded Person Re-Identification},'' \emph{IEEE Transactions on Neural
  Networks and Learning Systems}, vol.~33, no.~9, pp. 4624--4634, 2022.

\bibitem{wang2022review}
Z.~Wang, J.~Zhan, C.~Duan, X.~Guan, P.~Lu, and K.~Yang, ``{A Review of Vehicle
  Detection Techniques for Intelligent Vehicles},'' \emph{IEEE Transactions on
  Neural Networks and Learning Systems}, 2022.

\bibitem{zhang2015cross}
C.~Zhang, H.~Li, X.~Wang, and X.~Yang, ``{Cross-Scene Crowd Counting via Deep
  Convolutional Neural Networks},'' in \emph{Proceedings of the IEEE Conference
  on Computer Vision and Pattern Recognition}, 2015, pp. 833--841.

\bibitem{zhang2016single}
Y.~Zhang, D.~Zhou, S.~Chen, S.~Gao, and Y.~Ma, ``{Single-Image Crowd Counting
  via Multi-Column Convolutional Neural Network},'' in \emph{Proceedings of the
  IEEE Conference on Computer Vision and Pattern Recognition}, 2016, pp.
  589--597.

\bibitem{sindagi2017generating}
V.~A. Sindagi and V.~M. Patel, ``{Generating High-Quality Crowd Density Maps
  using Contextual Pyramid CNNs},'' in \emph{Proceedings of the IEEE
  International Conference on Computer Vision}, 2017, pp. 1861--1870.

\bibitem{babu2017switching}
D.~Babu~Sam, S.~Surya, and R.~Venkatesh~Babu, ``{Switching Convolutional Neural
  Network for Crowd Counting},'' in \emph{Proceedings of the IEEE Conference on
  Computer Vision and Pattern Recognition}, 2017, pp. 5744--5752.

\bibitem{boominathan2016crowdnet}
L.~Boominathan, S.~S. Kruthiventi, and R.~V. Babu, ``{CrowdNet: A Deep
  Convolutional Network for Dense Crowd Counting},'' in \emph{Proceedings of
  the 24th ACM International Conference on Multimedia}, 2016, pp. 640--644.

\bibitem{liu2019adcrowdnet}
N.~Liu, Y.~Long, C.~Zou, Q.~Niu, L.~Pan, and H.~Wu, ``{ADCrowdNet: An
  Attention-Injective Deformable Convolutional Network for Crowd
  Understanding},'' in \emph{Proceedings of the IEEE Conference on Computer
  Vision and Pattern Recognition}, 2019, pp. 3225--3234.

\bibitem{li2018csrnet}
Y.~Li, X.~Zhang, and D.~Chen, ``{CSRNet: Dilated Convolutional Neural Networks
  for Understanding the Highly Congested Scenes},'' in \emph{Proceedings of the
  IEEE Conference on Computer Vision and Pattern Recognition}, 2018, pp.
  1091--1100.

\bibitem{liu2019context}
W.~Liu, M.~Salzmann, and P.~Fua, ``{Context-Aware Crowd Counting},'' in
  \emph{Proceedings of the IEEE Conference on Computer Vision and Pattern
  Recognition}, 2019, pp. 5099--5108.

\bibitem{tian2019padnet}
Y.~Tian, Y.~Lei, J.~Zhang, and J.~Z. Wang, ``{PaDNet: Pan-Density Crowd
  Counting},'' \emph{IEEE Transactions on Image Processing}, vol.~29, no.~11,
  pp. 2714--2727, 2019.

\bibitem{cao2018scale}
X.~Cao, Z.~Wang, Y.~Zhao, and F.~Su, ``{Scale Aggregation Network for Accurate
  and Efficient Crowd Counting},'' in \emph{Proceedings of the European
  Conference on Computer Vision}, 2018, pp. 734--750.

\bibitem{chen2019scale}
X.~Chen, Y.~Bin, N.~Sang, and C.~Gao, ``{Scale Pyramid Network for Crowd
  Counting},'' in \emph{Proceedings of the IEEE Winter Conference on
  Applications of Computer Vision (WACV)}, 2019, pp. 1941--1950.

\bibitem{jiang2019learning}
X.~Jiang, L.~Zhang, P.~Lv, Y.~Guo, R.~Zhu, Y.~Li, Y.~Pang, X.~Li, B.~Zhou, and
  M.~Xu, ``{Learning Multi-Level Density Maps for Crowd Counting},'' \emph{IEEE
  Transactions on Neural Networks and Learning Systems}, vol.~31, no.~8, pp.
  2705--2715, 2020.

\bibitem{dong2022clrnet}
L.~Dong, H.~Zhang, J.~Ma, X.~Xu, Y.~Yang, and Q.~J. Wu, ``{CLRNet: A Cross
  Locality Relation Network for Crowd Counting in Videos},'' \emph{IEEE
  Transactions on Neural Networks and Learning Systems}, 2022.

\bibitem{gao2021domain}
J.~Gao, T.~Han, Y.~Yuan, and Q.~Wang, ``{Domain-adaptive Crowd Counting via
  High-quality Image Translation and Density Reconstruction},'' \emph{IEEE
  Transactions on Neural Networks and Learning Systems}, vol.~34, no.~8, pp.
  4803--4815, 2023.

\bibitem{zhou2021self}
C.~Zhou, C.~Xu, Z.~Cui, T.~Zhang, and J.~Yang, ``{Self-Teaching Video Object
  Segmentation},'' \emph{IEEE Transactions on Neural Networks and Learning
  Systems}, vol.~33, no.~4, pp. 1623--1637, 2022.

\bibitem{li2022self}
G.~Li, D.~Hong, K.~Xu, B.~Zhong, L.~Su, Z.~Han, and Q.~Huang, ``{Self
  Supervised Progressive Network for High Performance Video Object
  Segmentation},'' \emph{IEEE Transactions on Neural Networks and Learning
  Systems}, 2022.

\bibitem{xian2022location}
G.~Xian, C.~Ji, L.~Zhou, G.~Chen, J.~Zhang, B.~Li, X.~Xue, and J.~Pu,
  ``{Location-guided {LiDAR}-based Panoptic Segmentation for Autonomous
  Driving},'' \emph{IEEE Transactions on Intelligent Vehicles}, vol.~8, no.~2,
  pp. 1473--1483, 2023.

\bibitem{gu2022pixel}
{Gu, Zhangxuan and Zhou, Siyuan and Niu, Li and Zhao, Zihan and Zhang, Liqing},
  ``{From Pixel to Patch: Synthesize Context-Aware Features for Zero-Shot
  Semantic Segmentation},'' \emph{IEEE Transactions on Neural Networks and
  Learning Systems}, 2022.

\bibitem{ge2020cascaded}
S.~Ge, C.~Zhang, S.~Li, D.~Zeng, and D.~Tao, ``{Cascaded Correlation Refinement
  for Robust Deep Tracking},'' \emph{IEEE Transactions on Neural Networks and
  Learning Systems}, vol.~32, no.~3, pp. 1276--1288, 2020.

\bibitem{yu2022learning}
H.~Yu, P.~Zhu, K.~Zhang, Y.~Wang, S.~Zhao, L.~Wang, T.~Zhang, and Q.~Hu,
  ``{Learning Dynamic Compact Memory Embedding for Deformable Visual Object
  Tracking},'' \emph{IEEE Transactions on Neural Networks and Learning
  Systems}, 2022.

\bibitem{guo2023multi}
G.~Guo and S.~Zhao, ``{{3D} Multi-Object Tracking With Adaptive Cubature
  {K}alman Filter for Autonomous Driving},'' \emph{IEEE Transactions on
  Intelligent Vehicles}, vol.~8, no.~1, pp. 512--519, 2023.

\bibitem{predhumeau2023pedestrian}
M.~Prédhumeau, A.~Spalanzani, and J.~Dugdale, ``{Pedestrian Behavior in Shared
  Spaces with Autonomous Vehicles: An Integrated Framework and Review},''
  \emph{IEEE Transactions on Intelligent Vehicles}, vol.~8, no.~1, pp.
  438--457, 2023.

\bibitem{zhou2023multi}
W.~Zhou, S.~Dong, J.~Lei, and L.~Yu, ``{{MTAN}et: Multitask-aware Network with
  Hierarchical Multimodal Fusion for {RGB-T} Urban Scene Understanding},''
  \emph{IEEE Transactions on Intelligent Vehicles}, vol.~8, no.~1, pp. 48--58,
  2023.

\bibitem{guan2023autonomous}
M.~Guan and C.~Wen, ``{Autonomous Exploration Using {UWB} and {LiDAR}},''
  \emph{Journal of Automation and Intelligence}, vol.~2, no.~1, pp. 51--60,
  2023.

\bibitem{huang2023cost}
X.~Huang, Z.~Li, and F.~L. Lewis, ``{Cost-effective Distributed {FTFC} for
  Uncertain Nonholonomic Mobile Robot Fleet with Collision Avoidance and
  Connectivity Preservation},'' \emph{Journal of Automation and Intelligence},
  vol.~2, no.~1, pp. 42--50, 2023.

\bibitem{idrees2013multi}
H.~Idrees, I.~Saleemi, C.~Seibert, and M.~Shah, ``{Multi-Source Multi-Scale
  Counting in Extremely Dense Crowd Images},'' in \emph{Proceedings of the IEEE
  Conference on Computer Vision and Pattern Recognition}, 2013, pp. 2547--2554.

\bibitem{idrees2018composition}
H.~Idrees, M.~Tayyab, K.~Athrey, D.~Zhang, S.~Al-Maadeed, N.~Rajpoot, and
  M.~Shah, ``{Composition Loss for Counting, Density Map Estimation and
  Localization in Dense Crowds},'' in \emph{Proceedings of the European
  Conference on Computer Vision}, 2018, pp. 532--546.

\bibitem{wang2020nwpu}
Q.~Wang, J.~Gao, W.~Lin, and X.~Li, ``{NWPU-Crowd: A Large-Scale Benchmark for
  Crowd Counting and Localization},'' \emph{IEEE Transactions on Pattern
  Analysis and Machine Intelligence}, vol.~43, no.~6, pp. 2141--2149, 2020.

\bibitem{wang2019learning}
Q.~Wang, J.~Gao, W.~Lin, and Y.~Yuan, ``{Learning from Synthetic Data for Crowd
  Counting in the Wild},'' in \emph{Proceedings of the IEEE Conference on
  Computer Vision and Pattern Recognition}, 2019, pp. 8198--8207.

\bibitem{zhang2022game}
R.~Zhang, Q.~Zong, X.~Zhang, L.~Dou, and B.~Tian, ``{Game of Drones: Multi-UAV
  Pursuit-Evasion Game With Online Motion Planning by Deep Reinforcement
  Learning},'' \emph{IEEE Transactions on Neural Networks and Learning
  Systems}, 2022.

\bibitem{liu2018leveraging}
X.~Liu, J.~Van De~Weijer, and A.~D. Bagdanov, ``{Leveraging Unlabeled Data for
  Crowd Counting by Learning to Rank},'' in \emph{Proceedings of the IEEE
  Conference on Computer Vision and Pattern Recognition}, 2018, pp. 7661--7669.

\bibitem{sindagieccv2020learning}
V.~A. Sindagi, R.~Yasarla, D.~S. Babu, R.~V. Babu, and V.~M. Patel, ``{Learning
  to Count in the Crowd from Limited Labeled Data},'' in \emph{Proceedings of
  European Conference on Computer Vision}.\hskip 1em plus 0.5em minus
  0.4em\relax Springer, 2020, pp. 212--229.

\bibitem{yanliusemieccv2020}
Y.~Liu, L.~Liu, P.~Wang, P.~Zhang, and Y.~Lei, ``{Semi-Supervised Crowd
  Counting via Self-Training on Surrogate Tasks},'' in \emph{Proceedings of
  European Conference on Computer Vision}.\hskip 1em plus 0.5em minus
  0.4em\relax Springer, 2020, pp. 242--259.

\bibitem{liu2019exploiting}
X.~Liu, J.~Van De~Weijer, and A.~D. Bagdanov, ``{Exploiting Unlabeled Data in
  CNNs by Self-Supervised Learning to Rank},'' \emph{IEEE Transactions on
  Pattern Analysis and Machine Intelligence}, vol.~41, no.~8, pp. 1862--1878,
  2019.

\bibitem{zhang2020kgsnet}
Y.~Zhang, Y.~Bai, M.~Ding, S.~Xu, and B.~Ghanem, ``{KGSnet: Key-point-guided
  Super-resolution Network for Pedestrian Detection in the Wild},'' \emph{IEEE
  Transactions on Neural Networks and Learning Systems}, vol.~32, no.~5, pp.
  2251--2265, 2020.

\bibitem{dalal2005histograms}
N.~Dalal and B.~Triggs, ``{Histograms of Oriented Gradients for Human
  Detection},'' in \emph{Proceedings of the IEEE Conference on Computer Vision
  and Pattern Recognition}, vol.~1, 2005, pp. 886--893.

\bibitem{enzweiler2008monocular}
M.~Enzweiler and D.~M. Gavrila, ``{Monocular Pedestrian Detection: Survey and
  Experiments},'' \emph{IEEE Transactions on Pattern Analysis and Machine
  Intelligence}, vol.~31, no.~12, pp. 2179--2195, 2008.

\bibitem{tuzel2008pedestrian}
O.~Tuzel, F.~Porikli, and P.~Meer, ``{Pedestrian Detection via Classification
  on Riemannian manifolds},'' \emph{IEEE Transactions on Pattern Analysis and
  Machine Intelligence}, vol.~30, no.~10, pp. 1713--1727, 2008.

\bibitem{felzenszwalb2009object}
P.~F. Felzenszwalb, R.~B. Girshick, D.~McAllester, and D.~Ramanan, ``{Object
  Detection with Discriminatively Trained Part-Based Models},'' \emph{IEEE
  Transactions on Pattern Analysis and Machine Intelligence}, vol.~32, no.~9,
  pp. 1627--1645, 2009.

\bibitem{wu2005detection}
B.~Wu and R.~Nevatia, ``{Detection of Multiple, Partially Occluded Humans in a
  Single Image by Bayesian Combination of Edgelet Part Detectors},'' in
  \emph{Proceedings of the IEEE International Conference on Computer Vision},
  vol.~1, 2005, pp. 90--97.

\bibitem{chan2011counting}
A.~B. Chan and N.~Vasconcelos, ``{Counting People with Low-Level Features and
  Bayesian Regression},'' \emph{IEEE Transactions on image processing},
  vol.~21, no.~4, pp. 2160--2177, 2011.

\bibitem{lempitsky2010learning}
V.~Lempitsky and A.~Zisserman, ``{Learning to Count Objects in Images},'' in
  \emph{Advances in Neural Information Processing Systems}, 2010, pp.
  1324--1332.

\bibitem{wang2015deep}
C.~Wang, H.~Zhang, L.~Yang, S.~Liu, and X.~Cao, ``{Deep People Counting in
  Extremely Dense Crowds},'' in \emph{Proceedings of the 23rd ACM International
  Conference on Multimedia}, 2015, pp. 1299--1302.

\bibitem{chen2023crowd}
{Chen, Zhangping and Zhang, Shuo and Zheng, Xiaoqing and Zhao, Xiaodong and
  Kong, Yaguang}, ``{Crowd Counting Based on Multiscale Spatial Guided
  Perception Aggregation Network},'' \emph{IEEE Transactions on Neural Networks
  and Learning Systems}, 2023.

\bibitem{jiang2020attention}
X.~Jiang, L.~Zhang, M.~Xu, T.~Zhang, P.~Lv, B.~Zhou, X.~Yang, and Y.~Pang,
  ``{Attention Scaling for Crowd Counting},'' in \emph{Proceedings of the IEEE
  Conference on Computer Vision and Pattern Recognition}, 2020, pp. 4706--4715.

\bibitem{yang2020reverse}
Y.~Yang, G.~Li, Z.~Wu, L.~Su, Q.~Huang, and N.~Sebe, ``{Reverse Perspective
  Network for Perspective-Aware Object Counting},'' in \emph{Proceedings of the
  IEEE Conference on Computer Vision and Pattern Recognition}, 2020, pp.
  4374--4383.

\bibitem{wan2019adaptive}
J.~Wan and A.~Chan, ``{Adaptive Density Map Generation for Crowd Counting},''
  in \emph{Proceedings of the IEEE International Conference on Computer
  Vision}, 2019, pp. 1130--1139.

\bibitem{wan2020kernel}
J.~Wan, Q.~Wang, and A.~B. Chan, ``{Kernel-Based Density Map Generation for
  Dense Object Counting},'' \emph{IEEE Transactions on Pattern Analysis and
  Machine Intelligence}, vol.~44, no.~3, pp. 1357--1370, 2022.

\bibitem{zhu2023confusion}
{Zhu, Jiawen and Zhao, Wenda and Yao, Libo and He, You and Hu, Maodi and Zhang,
  Xiaoxing and Wang, Shuo and Li, Tao and Lu, Huchuan}, ``{Confusion Region
  Mining for Crowd Counting},'' \emph{IEEE Transactions on Neural Networks and
  Learning Systems}, 2023.

\bibitem{bai2020adaptive}
S.~Bai, Z.~He, Y.~Qiao, H.~Hu, W.~Wu, and J.~Yan, ``{Adaptive Dilated Network
  With Self-Correction Supervision for Counting},'' in \emph{Proceedings of the
  IEEE Conference on Computer Vision and Pattern Recognition}, 2020, pp.
  4594--4603.

\bibitem{shen2018crowd}
Z.~Shen, Y.~Xu, B.~Ni, M.~Wang, J.~Hu, and X.~Yang, ``{Crowd Counting via
  Adversarial Cross-Scale Consistency Pursuit},'' in \emph{Proceedings of the
  IEEE Conference on Computer Vision and Pattern Recognition}, 2018, pp.
  5245--5254.

\bibitem{yang2018multi}
J.~Yang, Y.~Zhou, and S.-Y. Kung, ``{Multi-Scale Generative Adversarial
  Networks for Crowd Counting},'' in \emph{Proceedings of the 24th
  International Conference on Pattern Recognition}, 2018, pp. 3244--3249.

\bibitem{sindagi2019multi}
V.~A. Sindagi and V.~M. Patel, ``{Multi-Level Bottom-Top and Top-Bottom Feature
  Fusion for Crowd Counting},'' in \emph{Proceedings of the IEEE International
  Conference on Computer Vision}, 2019, pp. 1002--1012.

\bibitem{xiong2019open}
H.~Xiong, H.~Lu, C.~Liu, L.~Liu, Z.~Cao, and C.~Shen, ``{From Open Set to
  Closed Set: Counting Objects by Spatial Divide-and-Conquer},'' in
  \emph{Proceedings of the IEEE International Conference on Computer Vision},
  2019, pp. 8362--8371.

\bibitem{lu2022counting}
H.~Lu, L.~Liu, H.~Wang, and Z.~Cao, ``{Counting Crowd by Weighing Counts: A
  Sequential Decision-Making Perspective},'' \emph{IEEE Transactions on Neural
  Networks and Learning Systems}, 2022.

\bibitem{xiao2023sampled}
B.~Xiao, H.-K. Lam, X.~Su, Z.~Wang, F.~P.-W. Lo, S.~Chen, and E.~Yeatman,
  ``{Sampled-data Control through Model-free Reinforcement Learning with
  Effective Experience Replay},'' \emph{Journal of Automation and
  Intelligence}, vol.~2, no.~1, pp. 20--30, 2023.

\bibitem{ma2019bayesian}
Z.~Ma, X.~Wei, X.~Hong, and Y.~Gong, ``{Bayesian Loss for Crowd Count
  Estimation with Point Supervision},'' in \emph{Proceedings of the IEEE
  International Conference on Computer Vision}, 2019, pp. 6142--6151.

\bibitem{lian2021locating}
D.~Lian, X.~Chen, J.~Li, W.~Luo, and S.~Gao, ``{Locating and Counting Heads in
  Crowds with A Depth Prior},'' \emph{IEEE Transactions on Pattern Analysis and
  Machine Intelligence}, vol.~44, no.~12, pp. 9056--9072, 2021.

\bibitem{lian2019density}
D.~Lian, J.~Li, J.~Zheng, W.~Luo, and S.~Gao, ``{Density Map Regression Guided
  Detection Network for {RGB-D} Crowd Counting and Localization},'' in
  \emph{Proceedings of the IEEE/CVF Conference on Computer Vision and Pattern
  Recognition}, 2019, pp. 1821--1830.

\bibitem{change2013semi}
C.~Change~Loy, S.~Gong, and T.~Xiang, ``{From Semi-Supervised to Transfer
  Counting of Crowds},'' in \emph{Proceedings of the IEEE International
  Conference on Computer Vision}, 2013, pp. 2256--2263.

\bibitem{wang2021neuron}
Q.~Wang, T.~Han, J.~Gao, and Y.~Yuan, ``{Neuron Linear Transformation: Modeling
  the Domain Shift for Crowd Counting},'' \emph{IEEE Transactions on Neural
  Networks and Learning Systems}, vol.~33, no.~8, pp. 3238--3250, 2022.

\bibitem{sam2019almost}
D.~B. Sam, N.~N. Sajjan, H.~Maurya, and R.~V. Babu, ``{Almost Unsupervised
  Learning for Dense Crowd Counting},'' in \emph{Proceedings of the AAAI
  Conference on Artificial Intelligence}, vol.~33, 2019, pp. 8868--8875.

\bibitem{liu2009learning}
T.-Y. Liu, ``{Learning to Rank for Information Retrieval},'' \emph{Foundations
  and Trends{\textregistered} in Information Retrieval}, vol.~3, no.~3, pp.
  225--331, 2009.

\bibitem{li2022learning}
H.~Li, \emph{{Learning to Rank for Information Retrieval and Natural Language
  Processing}}.\hskip 1em plus 0.5em minus 0.4em\relax Springer Nature, 2022.

\bibitem{ghanbari2022learning}
E.~Ghanbari and A.~Shakery, ``{A Learning to Rank Framework Based on
  Cross-lingual Loss Function for Cross-lingual Information Retrieval},''
  \emph{Applied Intelligence}, vol.~52, no.~3, pp. 3156--3174, 2022.

\bibitem{chen2023set}
X.~Chen, J.~Shen, W.~Xia, J.~Jin, Y.~Song, W.~Zhang, W.~Liu, M.~Zhu, R.~Tang,
  K.~Dong \emph{et~al.}, ``{Set-to-Sequence Ranking-based Concept-aware
  Learning Path Recommendation},'' \emph{arXiv preprint arXiv:2306.04234},
  2023.

\bibitem{zehlike2022fairness}
M.~Zehlike, K.~Yang, and J.~Stoyanovich, ``{Fairness in Ranking, Part {II}:
  Learning-to-Rank and Recommender Systems},'' \emph{ACM Computing Surveys},
  vol.~55, no.~6, pp. 1--41, 2022.

\bibitem{wang2023skellam}
H.~Wang, ``{Skellam Rank: Fair Learning to Rank Algorithm based on Poisson
  Process and Skellam Distribution for Recommender Systems},'' \emph{arXiv
  preprint arXiv:2306.06607}, 2023.

\bibitem{li2022confidence}
C.~Li, X.~Hu, and C.~Chen, ``{Confidence Estimation Using Unlabeled Data},'' in
  \emph{The Eleventh International Conference on Learning Representations},
  2022.

\bibitem{datta2008image}
R.~Datta, D.~Joshi, J.~Li, and J.~Z. Wang, ``{Image Retrieval: Ideas,
  Influences, and Trends of the New Age},'' \emph{ACM Computing Surveys},
  vol.~40, no.~2, pp. 1--60, 2008.

\bibitem{li2023gaitcotr}
J.~Li, Y.~Zhang, H.~Shan, and J.~Zhang, ``{Gait{COTR}: Improved
  Spatial-Temporal Representation for Gait Recognition with a Hybrid
  Convolution-Transformer Framework},'' in \emph{IEEE International Conference
  on Acoustics, Speech and Signal Processing}, 2023.

\bibitem{li2023motion}
J.~Li, J.~Gao, Y.~Zhang, H.~Shan, and J.~Zhang, ``Motion matters: A novel
  motion modeling for cross-view gait feature learning,'' in \emph{IEEE
  International Conference on Acoustics, Speech and Signal Processing}, 2023.

\bibitem{zagoruyko2015learning}
S.~Zagoruyko and N.~Komodakis, ``{Learning to Compare Image Patches via
  Convolutional Neural Networks},'' in \emph{Proceedings of the IEEE/CVF
  Conference on Computer Vision and Pattern Recognition}, 2015, pp. 4353--4361.

\bibitem{faigenbaum2022image}
S.~Faigenbaum-Golovin and O.~Shimshi, ``{Image Quality Assessment: Learning to
  Rank Image Distortion Level},'' \emph{arXiv preprint arXiv:2208.03317}, 2022.

\bibitem{liu2017rankiqa}
X.~Liu, J.~Van De~Weijer, and A.~D. Bagdanov, ``{Rank{IQA}: Learning from
  Rankings for No-Reference Image Quality Assessment},'' in \emph{Proceedings
  of the IEEE/CVF International Conference on Computer Vision}, 2017, pp.
  1040--1049.

\bibitem{simonyan2014vgg}
K.~Simonyan and A.~Zisserman, ``{Very Deep Convolutional Networks for
  Large-Scale Image Recognition},'' \emph{arXiv preprint arXiv:1409.1556},
  2014.

\bibitem{meng2021spatial}
Y.~Meng, H.~Zhang, Y.~Zhao, X.~Yang, X.~Qian, X.~Huang, and Y.~Zheng,
  ``{Spatial Uncertainty-Aware Semi-Supervised Crowd Counting},'' in
  \emph{Proceedings of the IEEE International Conference on Computer Vision},
  2021, pp. 15\,549--15\,559.

\bibitem{wang2023semi}
{Wang, Xin and Zhan, Yue and Zhao, Yang and Yang, Tangwen and Ruan, Qiuqi},
  ``{Semi-supervised Crowd Counting with Spatial Temporal Consistency and
  Pseudo-label Filter},'' \emph{IEEE Transactions on Circuits and Systems for
  Video Technology}, 2023.

\bibitem{wan2020modeling}
J.~Wan and A.~Chan, ``{Modeling Noisy Annotations for Crowd Counting},''
  \emph{Advances in Neural Information Processing Systems}, vol.~33, pp.
  3386--3396, 2020.

\bibitem{dai2023cross}
M.~Dai, Z.~Huang, J.~Gao, H.~Shan, and J.~Zhang, ``{Cross-Head Supervision for
  Crowd Counting with Noisy Annotations},'' in \emph{IEEE International
  Conference on Acoustics, Speech and Signal Processing}, 2023.

\end{thebibliography}

%



%




\end{document}